\def\year{2021}\relax
\documentclass[letterpaper]{article} 
\usepackage{aaai21}  
\usepackage{times}  
\usepackage{helvet} 
\usepackage{courier}  
\usepackage[hyphens]{url}  
\usepackage{graphicx} 
\usepackage{natbib}  
\usepackage{caption} 

\usepackage{array, makecell} %
\usepackage{listings}
\usepackage{amsmath}
\usepackage[switch]{lineno}
\usepackage{multirow}
\usepackage{xcolor}

\title{Sensitive Data Detection with High-Throughput Neural Network Models for Financial Institutions }
\author{
   Anh Truong\thanks{These authors have made equal contributions to this article.},
   Austin Walters\footnotemark[1],
   Jeremy Goodsitt\footnotemark[1]
   \\
}
\affiliations{
	Capital One\\
   Applied Research, Center for Machine Learning\\
   Anh.Truong@capitalone.com, Austin.Walters@capitalone.com, Jeremy.Goodsitt@capitalone.com\\
}

\begin{document}
\frenchspacing
\setlength{\belowcaptionskip}{-10pt}
\maketitle
\begin{abstract}
Named Entity Recognition has been extensively investigated in many fields. However, the application of sensitive entity detection for production systems in financial institutions has not been well explored due to the lack of publicly available, labeled datasets. In this paper, we use internal and synthetic datasets to evaluate various methods of detecting NPI (Nonpublic Personally Identifiable) information commonly found within financial institutions, in both unstructured and structured data formats. Character-level neural network models including CNN, LSTM, BiLSTM-CRF, and CNN-CRF are investigated on two prediction tasks: (i) entity detection on multiple data formats, and (ii) column-wise entity prediction on tabular datasets. We compare these models with other standard approaches on both real and synthetic data, with respect to F1-score, precision, recall, and throughput. The real datasets include internal structured data and public email data with manually tagged labels. Our experimental results show that the CNN model is simple yet effective with respect to accuracy and throughput and thus, is the most suitable candidate model to be deployed in the production environment(s). Finally, we provide several lessons learned on data limitations, data labelling and the intrinsic overlap of data entities.

\end{abstract}
\section{Introduction}
Named Entity Recognition (NER) is a subset of Natural Language Processing (NLP) used for identifying predefined entities in text. NER is used on both domain-specific text such as social network \cite{ner-social} and biomedical text extraction \cite{ner-bio1, ner-bio2}, as well as more general corpora \cite{ijcai2019-692, ner-dll3, kurniawan-louvan-2018-empirical}. With a few exceptions \cite{ner-finance}, the use of NER in finance has not been extensively studied as there do not exist publicly available, labeled datasets that contain the sensitive information. Within financial institutions, those sensitive information needs to be protected before data are uploaded on emails, githubs, or data repositories. In addition, there are often a large amount of unstructured and structured data with different formats that are stored across many internal storage systems. In this paper, we implement an NER system that overcomes the aforementioned obstacles. Our system generates data with sensitive information and consists of neural network models with both reasonably high accuracy and data throughput. These models have been optimized for architectures and data preprocessing over multiple computation resources. Our NER system concentrates on solving two problems: (i) predicting the presence of sensitive entities on different data formats, and (ii) predicting column-wise entities on tabular datasets in which each column contains only one type of entity. Below, we list our contributions in detail.

\textbf{Data generation with sensitive information.}
Due to the lack of public datasets with sensitive information, a collection of common sensitive entities is generated, from which we generate multiple datasets containing these entities under different formats. The generated datasets consist of  unstructured text with sensitive entities dispersed throughout, and structured single-column and multi-column data where each column contains a sensitive entity of the same type. Each structured data is represented in different file types, csv, json, and parquet. For evaluation purposes, we incorporate internal data and real email data extracted from spam email corpora.

\textbf{Neural network model optimization for production environments.}
We explored various experiments on multiple neural network models including CNN (Convolutional Neural Network), LSTM (Long Short-Term Memory), CNN-LSTM, and CRF (Conditional Random Field) based models, CNN-CRF, BiLSTM-CRF, where CRF layer is utilized as a tag decoder.  We also evaluated the popular NER library SpaCy with its default NER configurations and retrained on our training datasets. Finally, for a baseline comparison, a regex model including a set of hand curated regex filters designed specifically for the entities in our dataset, as well as a CRF model with handcrafted features are used.

We observe that our CNN model outperformed the other models  with respect to both accuracy and throughput, and thus is the most suitable model for production usage. This model is also universal: it easily adapts from the entity detection task to column-wise entity prediction on tabular datasets with some slight modifications on input data preprocessing, and it is comparable to a specifically designed model for this task, Column-CNN-CRF.
\section{Related Work}
Sensitive information detection in structured text has been explored in different fields such as military and politics \cite{sensitive-politic-military}, and social network \cite{sensitive-social}. However, their work focused on sensitive document classification problems as opposed to our NER problem which predicts sensitive entities in text, under the financial domain. Approaches for NER include statistical modeling methods \cite{hmm, memm, crf} and recently neural network models \cite{ner-dll5, ner-dll6, ner-dll7}. In this paper, we seek for effective NER models which obtain not only high accuracy but also high throughput. To that end, we first optimize two models in this framework, CNN and LSTM \cite{ner-dll1}, evaluating different architectures to extract richer features LSTM-CNN \cite{lstm-cnn}. We then examine the effect of the CRF layer as a tag decoder to optimize the label sequence prediction. This has been observed to be an effective addition by \cite{cnn-crf} with CNN-CRF, and \cite{ner-dll1, ner-finance} with BiLSTM-CRF. It is worth noting that the aforementioned models are based on word embeddings, or the combination of word embeddings and character embeddings. In this work, we focus on character-level models as opposed to word-level models as the considered sensitive entities, specific to the financial domain, cannot be represented by the existing tokenizers trained on the general corpora \cite{bert-char}. Moreover, the sensitive entities we are trying to identify are unique and non-public (NPI), meaning that word-level representation of these entities potentially results in being out-of-vocabulary.  That being said, later made model comparisons do include a word-level NER model from the standard NLP toolkit, spaCy which are congruent with \cite{ner-dll3}, showing its disadvantage with such types of data.  Beside spaCy, there exists other well-known NER tools such as NLTK and recently Presidio from Microsoft, a tool for PII detection on text and image. However, spaCy was selected for comparison as it is ubiquitous and has been utilized in many studies \cite{spacy1, spacy2, spacy4}.
\section{Sensitive Data Generation}
The datasets used for model training and evaluation are created via synthetic generation and labeling since a public dataset for the desired entities did not exist. These datasets contain both sensitive and non-sensitive entities which would commonly be found within a financial institution's database. 
\subsection{Sensitive Entity Generation}
Below is the list of rules associated with the 19 entities considered in this paper (the detailed examples of these entities are given in the Appendix):\\\\
\textbf{Sensitive entities}:\\
\textbf{Address (ADDRESS)}: US address which may be multi-line and can contain newline characters. "City/City, State" are not included in this set.\\
\textbf{Bank Account Number (BAN)}: Numbers associated with an account at a bank. While technically these could be alphanumeric 1-18 in length, we limited to 10-18 digits as those are the far more common.\\
\textbf{Credit Card Number (CREDIT\_CARD)}: Amex, Visa, Mastercard, and Discover credit card numbers with the optional delimiters (",", " ", ".", "\_").\\
\textbf{Datetime (DATETIME)}: Datetime formats recognizable by the python datetime library.\\
\textbf{Email Address (EMAIL\_ADDRESS)}: An email address or email portion of the URI not contained within a url.\\
\textbf{Hash or Encryption Key (HASH\_OR\_KEY)}: Randomized concatenation of 16 or more alphanumeric characters and special characters "=", "-", "\textbackslash ", "/", "+" for md5, SHA1, SHA256, SHA512, or any encryption key.\\
\textbf{IPv4 (IPV4)}: Standard IPv4 formats.\\
\textbf{IPv6 (IPV6)}: Standard IPv6 formats.\\
\textbf{MAC Address (MAC\_ADDRESS)}: Standard MAC Address formats with delimiters ("-", ":", " ", ".").\\
\textbf{Person’s Name (PERSON)}: Name of a person including titles, but not including possessive nouns.\\
\textbf{Phone Number (PHONE\_NUMBER)}: US/Canada/UK/Ireland phone numbers with or without international code, zip code, or extensions.\\
\textbf{Social Security Number (SSN)}: US Social Security Numbers with varying delimiters ("-", ".", "", " ").\\
\textbf{URL (URL)}: Any url with a network scheme, www initialized, or valid url if the aforementioned were prepended.\\
\textbf{UUID (UUID)}: UUID4 format.\\\\
\textbf{Non-Sensitive Entities}:\\
\textbf{Background (BACKGROUND)}: Any text which does not fall into the other categories.\\
\textbf{Float (FLOAT)}: Digits with single decimal surrounded by whitespace or punctuation.\\
\textbf{Integer (INTEGER)}: Non-sensitive digits without decimal, surrounded by whitespace or punctuation.\\
\textbf{Ordinal (ORDINAL)}: Representations of order/position via words or alphanumeric mixes or version identifiers. Versions must have an alpha character associated with the version unless it is 3 digits separated by periods.\\
\textbf{Quantity (QUANTITY)}: Integer or float prepended/appended with text or quantity characters.

Throughout the paper, we refer \textit{entity/entity type} as entity category, \textit{entity format} as variants per entity type, and \textit{entity value} as values per format of each entity. For example, entity CREDIT\_CARD has a format XXXX-XXXX-XXXX-XXXX, of which one of the values is 1111-2222-3333-4444. 
\subsection{Synthetic Data Generation}
Both structured and unstructured data formats are synthetically generated: (1) unstructured text, (2) multi-column structured data, and (3) single-column structured data.
\subsubsection{ Unstructured Text}
Unstructured lines of text are randomly generated without context for entity placement. To incorporate different text structures, the following text formats are included. (a) sentences, (b) JSON, and (c) delimited data. JSON and delimited data are ingested as text to investigate the case where such structured data are read as text intentionally or unintentionally due to the schema error. For all formats, a list of adjectives, adverbs, nouns, verbs, stop words, and punctuation are generated from the WordNet corpus \footnote{https://wordnet.princeton.edu/} and considered background words. These words occur with a probability distribution obtained from the brown corpus \footnote{http://korpus.uib.no/icame/brown/bcm.html}, which are then randomly uppercased and combined with the random delimiters such as ``-" and ``,". For a non-background word, the selection is uniform, random across all other entities. General tunable parameters are given in the Appendix. Combining background and NPI words, the three unstructured formats are generated as follows:
\begin{itemize}
\item \textit{ Sentences}: Data are generated by first randomly determining the number of background words in a sentence, and then iteratively, randomly determining if each subsequent word is background or otherwise until the number number of background words is met.
\item \textit{ JSON}: Words are randomly generated for keys and values with no allowed nesting.
\item \textit{ Delimited Data}: Similar to random sentences, words are iteratively generated except separated by a randomly chosen delimiter (",",   " ", "\textbackslash t", ";", "\textbackslash x00", "\textbackslash x01").
\end{itemize}
The above formats can be tuned to have different lengths and probability of NPI occurrence or co-occurrence.
\subsubsection{ Multi-Column Structured Data}
Tabular datasets with 1 to 20 columns are generated with random schemas in CSV, PARQUET, JSON, and AVRO formats. Each column has a 50\% probability of being background with uniform random probability for all other entities. All entities within a column contain the same format.
\subsubsection{ Single-Column Structured Data}
Tabular single-column datasets are generated in CSV, PARQUET, JSON, and AVRO formats. Each dataset has uniform random probability for all entities. All entities within the dataset contain the same format.
\subsection{Internal Structured Data}
Twenty-five schemas found within a financial institution were replicated via synthetic data, matching the format and statistics of the underlying real data. Each dataset contains 1000 samples (where a row is a sample). When the dataset has a tabular schema, CSV, PARQUET, AVRO, and JSON files are created. Otherwise, only JSON and AVRO datasets are created due to the nested structure of the dataset.
\subsection{Public Email}
A random subsample of emails from the Enron, Trec07, and ADCG SS14 Challenge corpora \footnote{https://www.cs.cmu.edu/~enron/, https://plg.uwaterloo.ca/~gvcormac/treccorpus07/, https://www.kaggle.com/c/adcg-ss14-challenge-02-spam-mails-detection/data} are selected for manual labeling. For each corpus, 200 unique emails are randomly selected. Each email can have full header format or reduced header format containing only Date, From, To, and Subject. Not all entities existed in this dataset.
\subsection{Datasets for Entity Detection Task}
Datasets and the corresponding number of entities for train and test data are given in Table \ref{table:train-test-data}. For synthetic structured data, the train (test) data contain 50 (30) variations of multi-column schemas each with 100 (50) samples, and 250 (25) variations of single-column schemas each with 200 (200) samples. The tunable parameters for unstructured text are detailed in Table \ref{table:traindata-unstructure} of the Appendix.
\begin{table}[thb]
\scriptsize
\centering\renewcommand\cellalign{lc}
\setcellgapes{3pt}\makegapedcells
\begin{tabular}{|l|l|l|}
\hline
\textbf{Data}   &  \textbf{Train} &  \textbf{Test} \\
\hline
\textbf{Total (K)} & 413 & 841\\
\textbf{Unstructured Text (K)} & 193 & 22\\
\textbf{Multi-column Structured  Data (K)} & 121 & 19\\
\textbf{Single-column Structured Data (K)} & 99 & 32\\
\textbf{Internal Structured Data (K)} & \hspace{0.2cm} & 25\\
\textbf{Public Email (K)} & \hspace{0.2cm} & 768\\
\hline
\end{tabular}
\caption{Number of entities in the train and test datasets}
\label{table:train-test-data}
\end{table}
\subsection{Datasets for Column-wise Entity Prediction Task}
Separate training and testing datasets are generated for evaluation of columnar-level models, which predict entity type for a given column.
\subsubsection{Training}
Each training dataset contains 75k entity values. To investigate the value-grouping effect on the prediction accuracy, values are randomly subsampled for a given entity type and aggregated in different sizes. Ten datasets are generated with aggregate sizes ranging from 1 to 10.
\subsubsection{Testing}
Testing datasets are obtained from synthetic structured testing datasets and internal structured testing datasets. As with the training dataset, values are randomly subsampled from a column in a given dataset and aggregated with sizes ranging from 1 to 10. This aggregation approach is resampled 10 times for the columns in each dataset.
\section{Modeling Approaches}
\subsection{Character-level Neural Network Models}
\begin{figure}[!b]
\begin{center}
\centerline{\includegraphics[width=0.65\columnwidth]{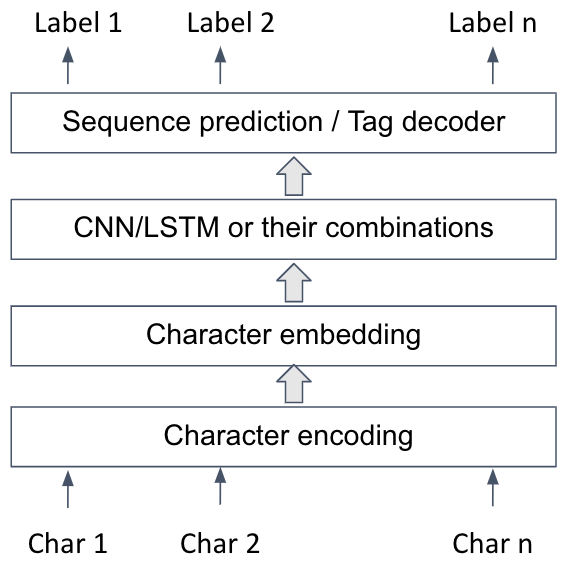}}
\caption{Character-level neural network model architecture}
\label{fig:model-generic}
\end{center}
\end{figure}
\begin{figure}[!tb]
\begin{center}
\centerline{\includegraphics[width=0.99\columnwidth]{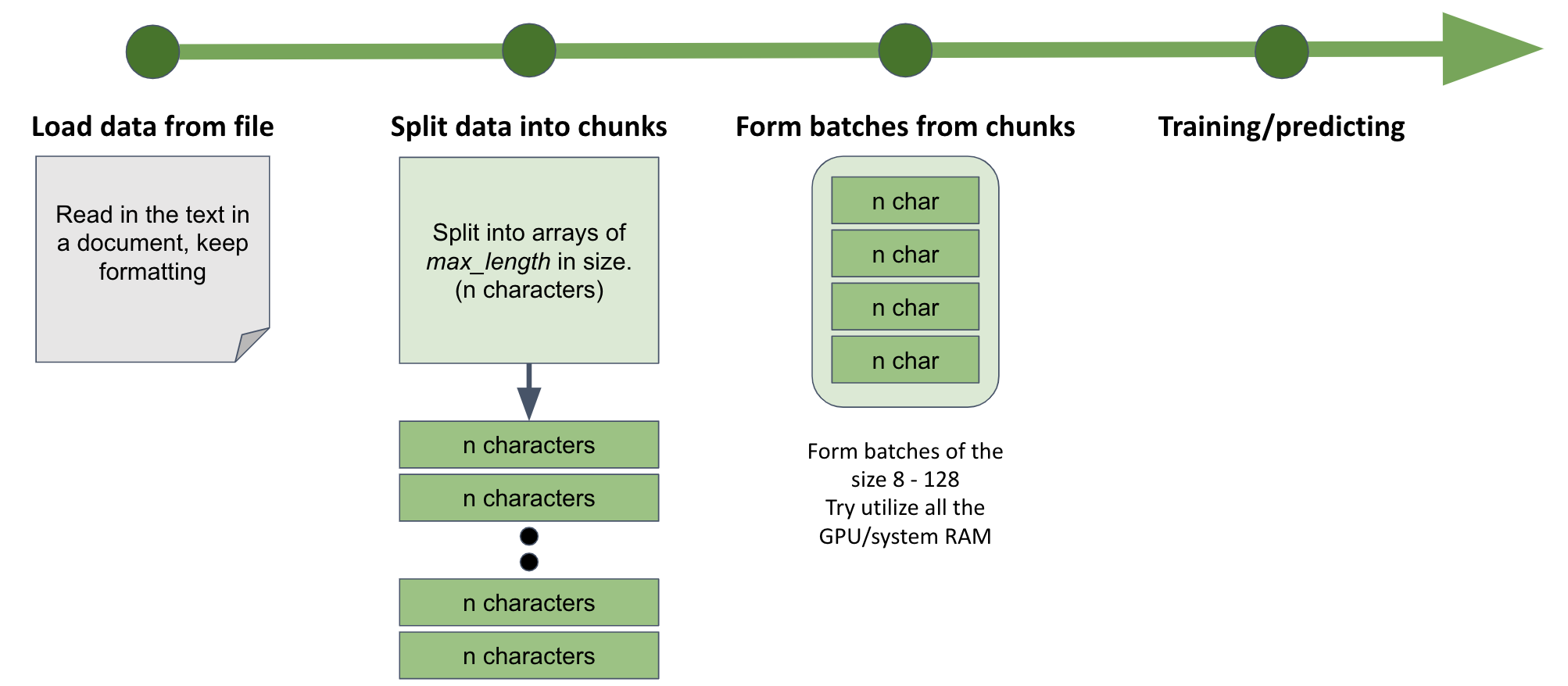}}
\caption{Flattening preprocessing for input text}
\label{fig:flattening}
\end{center}
\end{figure}

Our general character-level model architecture is given in Figure \ref{fig:model-generic}. Input strings are character encoded using ASCII indices. As string lengths vary across samples, the zero padding is applied at the end of each sample. To represent these encoded indices in latent character features, a pretrained Glove Character embedding \cite{glove} is used. The embeddings are fed into the next layers such as CNN and LSTM to extract more detailed features. Finally, a prediction network, fully connected layers or a tag decoder such as CRF, is applied to optimize the sequence prediction. From this general model architecture, the following models are considered in our paper: CNN, LSTM, CNN-LSTM, CNN-CRF, and BiLSTM-CRF. These models are selected to investigate (i) the effect of convolutional layer, recurrent network layer and their combination on feature learning, and (ii) the effect of additional tag decoder layer (CRF) to the overall prediction accuracy of the models.\\\\
\textbf{Data processing optimizations}\\
In order to maximize the throughput of our models in a production environment, the input character encoding component is integrated into the tensorflow computation graph. In addition, a flattening mechanism is applied to the input data as illustrated in Figure \ref{fig:flattening}. Input characters are all concatenated and chunked into an array of \textit{max\_length} characters. This step increases the model throughput on both CPU instances (6x) and GPU instances (3x-4x).
\begin{figure*}[!htb]
\begin{center}
\centerline{\includegraphics[width=1.99\columnwidth]{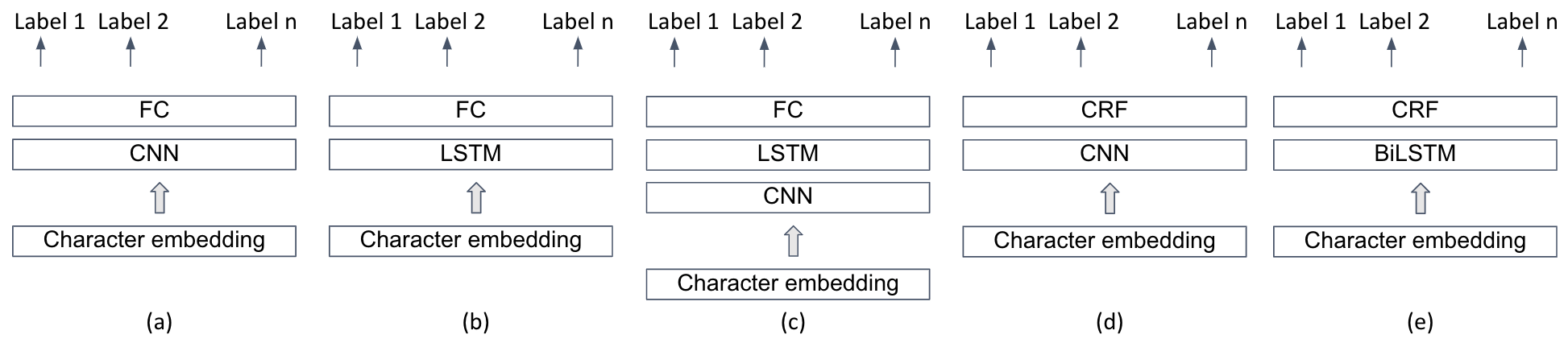}}
\caption{Character-level models. (a) Character-level CNN, (b) Character-level LSTM, (c) Character-level CNN-LSTM, (d) Character-level CNN-CRF, (e) Character-level BiLSTM-CRF}
\label{fig:model-char-level}
\end{center}
\end{figure*}
\subsection{CNN Model}
Detailed in Figure \ref{fig:model-char-level}(a), the core components of the CNN model are the four convolutional blocks followed by two connected blocks.
Each convolutional block consists of a 1-d convolutional layer, dropout layer, and batch normalization layer. Each fully-connected block consists of a dense layer and dropout layer. Dropout layers help regularize the network, and increases the accuracy since our data contains random context. Batch normalization is applied to reduce the effect of the internal covariate shift \cite{batchnorm}. Through manual optimization, the CNN model obtains the best accuracy with the following parameters: \textit{epochs=10, num-conv-layer=4, num-dense-layer=2, batch-size=24, embedding-dimension=64, max-length=3400, filter-size=13, dense-layer-size=96, and dropout=7.3\%}.
\subsection{LSTM Model}
Throughout manual optimization, only one LSTM layer is utilized and illustrated in Figure \ref{fig:model-char-level}(b). The best parameters for the LSTM model are given as: \textit{epochs=10, num-lstm-layer=1, batch-size=24, embedding-dimension=64, max-length=3400, lstm-size=64, activation=tanh, recurrent-dropout=10\%, dense-layer-size 32, and dropout 10\%}. CuDNNLSTM layers are used instead of the LSTM layer to optimize throughput as we will see later that CuDNNLSTM shows the significant run-time  improvement without sacrificing the accuracy.
\subsection{CNN-LSTM Model}
We consider the combined model in which the CNN layers as in Figure \ref{fig:model-char-level}(a) is put before the LSTM layer followed by the 2-layer dense network. The detailed model is given in Figure \ref{fig:model-char-level}(c).
Similar to the LSTM model, CuDNNLSTM layer is used in place of LSTM layer to optimize throughput. The best obtained parameters are given as: \textit{epochs=10, num-conv-layer=4, num-lstm-layer=1, num-dense-layer=2, batch-size=24, embedding-dimension=64, max-length=3400, filter-size=13, lstm-size=64, activation=tanh, recurrent-dropout=10\%, dense-layer-size=96, and dropout=10\%}.
\subsection{CRF-based Models}
CRF has shown provable advantages over the fully connected layer for tag decoding step as it is able to learn the label of each character based on its neighbors \cite{ner-dll1, ner-dll3}. To investigate the effectiveness of this CRF layer as tag decoder, two following models are investigated.

CNN-CRF: same architecture as the CNN model except the dense layers are replaced by the CRF layer illustrated in Figure \ref{fig:model-char-level}(d). Similar to LSTM related models, CRF based models run slowly compared to other models. Through the optimization process, the best parameters for this model is given as: \textit{epochs=15, num-crf-layer=1, num-conv-layer=4, batch-size=128, embedding-dimension=64, max-length=3400, filter-size=13, dropout=7.3\%, optimizer=rmsprop}.

BiLSTM-CRF: suggested as the best model on various NER tasks \cite{ner-dll1}, the BiLSTM layer is swapped with the CNN layer from the CNN-CRF model as depicted in Figure \ref{fig:model-char-level}(e). The best obtained parameters are as follows: \textit{epochs=8, num-crf-layer=1, num-bilstm-layer=1, batch-size=128, embedding-dimension=64, max-length=2500, bilstm-layer=1, lstm-size=64, activation=tanh, recurrent-dropout=0\%, merge-mode=concat, dropout=20\%, optimizer=rmsprop}.
\subsection{Existing NER Models}
\subsubsection{Regex}
A list of regular expressions for all entities except PERSON (too large of a search space) and BACKGROUND is manually generated with respect to the training dataset. All regex rules are applied to the input text, sets of characters not matching a rule are considered BACKGROUND and labels were evenly split for ties. Since regex can become quite complex, only simple regex expressions or those which are quickly discoverable online are used. Additionally, each regex pattern is surrounded by encapsulators which ensure that any matching string in unstructured text is delimited by the specified characters and are not part of the match itself. The detailed regex pattern is given in the Appendix.
\subsubsection{CRF Model with Handcrafted Features (Ngram-CRF)}
A standalone CRF model with the following handcrafted feature extraction is considered:\\\\
-\textit{char.lower()}: get the lowercase character\\
-\textit{char.isupper()}: check if the character is uppercased\\
-\textit{char.isdigit()}: check if the character is digit\\
-\textit{char.isalnum()}: check if the character is alphanumeric\\\\
For each character, its extracted features are combined with features of its neighbors within a sliding window, instead of being fed into encoding and embedding components. Through manual optimization, the model obtain the best results with the parameters: \textit{window-len=4, batch-size=1000, max-length=2500, l1-coefficient=0.1, l2-coefficient=0.1, max-iterations=100, all-possible-transitions=True, all-possible-states=True}
\subsubsection{spaCy Model}
The spaCy model is fine-tuned on our training datasets with the default parameters. As spaCy processes data at the token level, the input strings are fed to this model without splitting to the character level. However, at the evaluation stage, the predictions from the spaCy model are split into character labels from token results to compare with other models.
\subsection{Columnar-Level Models}
Four model variations are evaluated:  (1) the best character-level model trained on the unstructured training dataset given in Table \ref{table:train-test-data} (Char-Best-Pretrained), (2) the best character-level model trained on the columnar-level dataset (Char-Best-Retrained), (3) a columnar-based CNN-CRF model (Column-CNN-CRF), and (4) a columnar-based CNN-BiLSTM model (Column-CNN-BiLSTM). Each model utilizes subsampled columns to make generalized predictions for an entire column.
\subsubsection{Best Character-Level Model}
For the best character-level models, data are preprocessed by taking a number of sampled rows per column and concatenating  them into a single sample delimited by five \textbackslash x01 characters.
Postprocessing on the model output is applied to convert the character entity values into a single subsample entity by taking the mode of character entity values, excluding PAD and separator characters. In cases of a tie during prediction, a non-background entity is randomly selected.
\subsubsection{Columnar-Based CNN Models}
\begin{figure}[!htb]
\begin{center}
\centerline{\includegraphics[width=0.7\columnwidth]{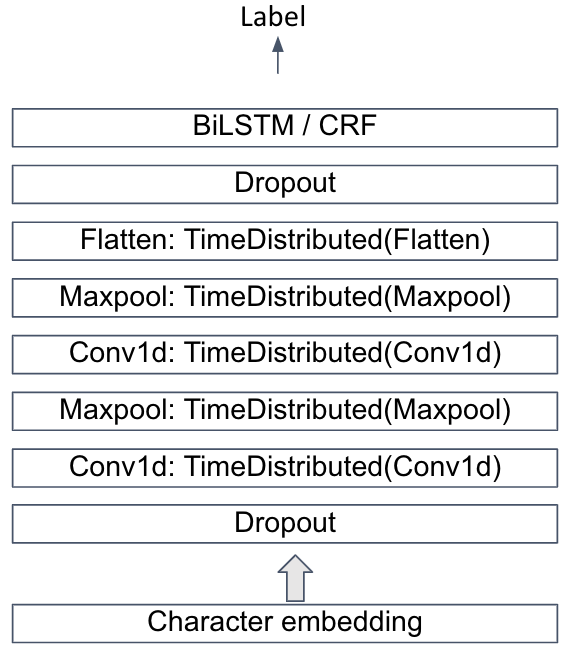}}
\caption{Columnar-level neural network model}
\label{fig:model-structured}
\end{center}
\end{figure}
\begin{figure}[!htb]
\begin{center}
\centerline{\includegraphics[width=0.95\columnwidth]{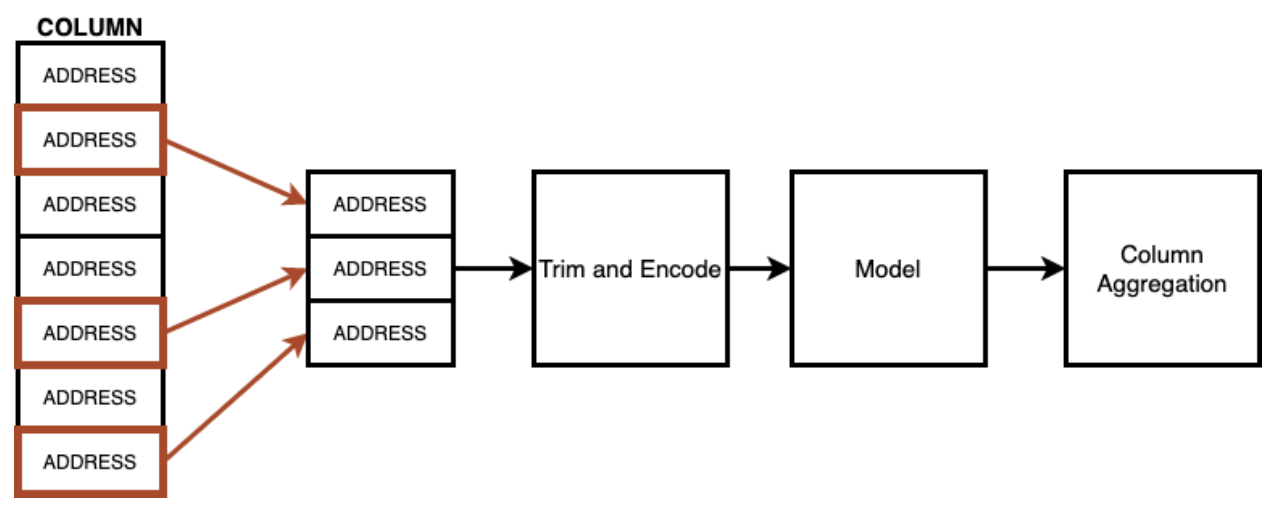}}
\caption{Columnar-level model workflow}
\label{fig:model-preprocess-structured}
\end{center}
\end{figure}
Subsampled rows in a column were processed similarly to words in a sentence  \cite{lstm-cnn, ner-bio2} within the columnar-based CNN approaches and fed into the model as a single sample. Before being fed into the model, each “word” is limited to 52 characters and subsequently encoded. The model architecture, derived from \cite{lstm-cnn}, is described in Figure \ref{fig:model-structured}.
The model output is an entity per subsampled row within the column which was aggregated via the mode into a single value, identical to the best character-level models. The following were the parameter values for these models (where applicable):  \textit{dropout=0.5, num-conv-filters=30, conv-size=9, dropout-recurrent=0.25, lstm-state-size=100, learning-rate=0.0105, optimizer='nadam', num-conv-layers=2, embedding-dim=30}.
Postprocessing on the model output is applied to convert the “word” entity values into a single subsample entity by taking the mode of character entity values, as seen in Figure \ref{fig:model-preprocess-structured}. In cases of a tie during prediction, a non-background entity is randomly selected. This subsample entity is the assumed generalized entity selection for the column.
\begin{table*}[thb]
\scriptsize
\centering\renewcommand\cellalign{lc}
\setcellgapes{3pt}\makegapedcells
\begin{tabular}{|l|l|l|l|l|l|}
\hline
\textbf{Model / Datasets}   & \makecell{\textbf{Multi-column,} \\ \textbf{Structured Data}}   & \makecell{\textbf{Single-column,} \\ \textbf{ Structured Data}}   & \makecell{\textbf{Unstructured Text} \\ \textbf{}}   & \makecell{\textbf{Public Emails} \\ \textbf{}}   & \makecell{\textbf{Internal,} \\ \textbf{Structured  Data}}\\
\hline
Char CNN & \makecell{\textbf{(0.99, 0.99, 0.99)} \\ \textbf{(0.98, 0.97, 0.97)}} & \makecell{\textbf{(0.99, 0.99, 0.99)} \\ \textbf{(0.97, 0.97, 0.97)}} & \makecell{\textbf{(0.98, 0.96, 0.97)} \\ \textbf{(0.96, 0.93, 0.94)}} & \makecell{(0.67, 0.79, 0.73) \\  (0.54, 0.78, 0.60)} & \makecell{(0.82, 0.87, 0.84) \\ (0.74, 0.83, 0.72)} \\
\hline
Char CuDNNLSTM  & \makecell{(0.94, 0.93, 0.93) \\ (0.89, 0.89, 0.88)} & \makecell{(0.95, 0.93, 0.94) \\ (0.87, 0.82, 0.84)} & \makecell{(0.84, 0.83, 0.83) \\ (0.82, 0.76, 0.76)} & \makecell{(0.55, 0.61, 0.58) \\ (0.46, 0.64, 0.51)} & \makecell{(0.54, 0.80, 0.65) \\ (0.54, 0.70, 0.59)} \\
\hline
Char CNN + CuDNNLSTM & \makecell{\textbf{(0.99, 0.98, 0.99)} \\ \textbf{(0.97, 0.96, 0.97)}} & \makecell{\textbf{(0.98, 0.98, 0.98)} \\ (0.93, 0.93, 0.93)} & \makecell{(0.96, 0.94, 0.95) \\ (0.92, 0.89, 0.89)} & \makecell{\textbf{(0.73, 0.75, 0.74)} \\ \textbf{(0.61, 0.75, 0.64)} } & \makecell{\textbf{(0.84, 0.90, 0.87)} \\ \textbf{(0.76, 0.85, 0.75)}} \\
\hline
Word spaCy  & \makecell{(0.86, 0.84, 0.85) \\ (0.82, 0.74, 0.77)} & \makecell{(0.96, 0.96, 0.96) \\ (0.93, 0.90, 0.90)} & \makecell{(0.75, 0.77, 0.76) \\ (0.74, 0.70, 0.71)} & \makecell{(0.62, 0.62, 0.62) \\ (0.52, 0.60, 0.50)} & \makecell{(0.48, 0.66, 0.56) \\ (0.53, 0.67, 0.48)} \\
\hline
Char Ngram + CRF & \makecell{(0.97, 0.94, 0.95) \\ (0.90, 0.87, 0.88)} & \makecell{(0.97, 0.96, 0.96) \\ (0.91, 0.85, 0.86)} & \makecell{(0.91, 0.88, 0.90) \\ (0.82, 0.80, 0.80)} & \makecell{(0.64, 0.71, 0.67) \\ (0.49, 0.67, 0.54)} & \makecell{(0.74, 0.84, 0.79) \\ (0.67, 0.74, 0.68)} \\
\hline
Char CNN + CRF & \makecell{\textbf{(0.99, 0.99, 0.99)} \\ \textbf{(0.98, 0.98, 0.98)}} & \makecell{\textbf{(0.99, 0.99, 0.99)} \\ \textbf{(0.96, 0.96, 0.96)}} & \makecell{\textbf{(0.98, 0.97, 0.97)} \\ \textbf{(0.96, 0.94, 0.95)}} & \makecell{\textbf{(0.70, 0.81, 0.75)} \\ \textbf{(0.55, 0.79, 0.62)}} & \makecell{\textbf{(0.83, 0.92, 0.87)} \\ \textbf{(0.71, 0.85, 0.74)}}\\
\hline
Char BiLSTM + CRF & \makecell{\textbf{(0.99, 0.98, 0.99)} \\ \textbf{(0.98, 0.96, 0.97)}} & \makecell{\textbf{(0.99, 0.98, 0.98)} \\ (0.97, 0.93, 0.95)} & \makecell{\textbf{(0.98, 0.95, 0.96)} \\ (0.95, 0.92, 0.93)} & \makecell{\textbf{(0.71, 0.78, 0.74)} \\ (0.56, 0.73, 0.60)} & \makecell{(0.79, 0.86, 0.82) \\ (0.73, 0.81, 0.71)} \\
\hline
Char Regex & \makecell{(0.71, 0.73, 0.72) \\ (0.76, 0.60, 0.63)} & \makecell{(0.91, 0.81, 0.85) \\ (0.79, 0.55, 0.61)} & \makecell{(0.67, 0.67, 0.67) \\ (0.73, 0.58, 0.60)} & \makecell{(0.64, 0.62, 0.63) \\ (0.54, 0.62, 0.53)} & \makecell{(0.31, 0.50, 0.38) \\ (0.61, 0.54, 0.49)}\\
\hline
\end{tabular}
\caption{Evaluation results on the test set for sensitive entities detection. In each cell, the first line and second line shows the micro and macro average results, respectively. Each line represents precision, recall and F1-score respectively. Synthetic and real data are given in the first three columns and the last two columns, respectively.}
\label{table:accuracy-sensitive}
\end{table*}
\begin{table}[thb]
\scriptsize
\centering\renewcommand\cellalign{l}
\setcellgapes{3pt}\makegapedcells
\begin{tabular}{|p{2.7cm}|p{0.8cm}|p{0.9cm}|p{1cm}|p{0.8cm}|}
\hline
{\tiny \textbf{Model / EC2 Instance Type}}   & {\tiny \textbf{c5.2xlarge}} & {\tiny \textbf{g4dn.xlarge}} & {\tiny \textbf{g4dn.8xlarge}} & {\tiny \textbf{p3.2xlarge}}\\
\hline
Char CNN                 & \textbf{3.01}       & \textbf{18.08}       & \textbf{18.18}        & \textbf{28.53}      \\
\hline
Char LSTM                & 0.9574     & 0.4668      & 0.4328       & 0.3833     \\
\hline
Char CuDNNLSTM           & N/A        & 7.893       & 7.102        & 6.346      \\
\hline
Char CNN+LSTM            & 0.7675     & 0.4407      & 0.4025       & 0.3701     \\
\hline
Char CNN+CuDNNLSTM       & N/A        & 4.704       & 4.258        & 5.031      \\
\hline
Word spaCy (Unflattened) & 0.054      & 0.045       & 0.045        & 0.032      \\
\hline
Word spaCy (Flattened)   & 0.071      & 0.134       & 0.13         & 0.11       \\
\hline
Char Ngram + CRF         & 0.087      & 0.071       & 0.087        & 0.087      \\
\hline
Char CNN + CRF           & 0.8673     & 0.312       & 0.4014       & 0.2455     \\
\hline
Char BiLSTM + CRF        & 0.543      & 0.0839      & 0.0979       & 0.0685     \\
\hline
Char Regex               & 0.8388     & 0.7344      & 0.734        & 0.6055     \\
\hline
\end{tabular}
\caption{Throughput in GB/hr evaluation on four different AWS EC2 instances}
\label{table:throughput-unstructured}
\end{table}
\section{Evaluation Results}
\subsection{Evaluation Metrics}
In this paper, precision, recall, and F1-score are used as the model accuracy comparison metrics. Performance on the test dataset is evaluated using micro and macro averages across all entities excluding PAD and BACKGROUND. In addition to accuracy, model throughput, measured as GB of data processed per hour (GB/hr), is evaluated on both CPU and GPU AWS EC2 instances.
\subsection{Entity detection on Multiple Data Formats}
\subsubsection{Accuracy Evaluation}
Table \ref{table:accuracy-sensitive} shows results of the models on the test set. Model performance is consistently higher on the synthetic datasets which have more similar schema and context to the training dataset as opposed to the internal and email datasets which do not. Additionally, regex, spaCy, Ngram-CRF, and the LSTM model are less accurate than the other models. Models with CRF layer as the tag decoder such as CNN-CRF, BiLSTM-CRF, and the combined CNN-CuDNNLSTM model provide marginal to no improvement over the CNN model. Detailed results regarding individual entities are described in the Appendix.
\subsubsection{Throughput Evaluation}
Model throughput is evaluated on a CPU instance, c5.2xlarge (8 vCPU, 16 GiB Memory), and varying GPU instances, g4dn.xlarge (4 vCPU, 16 GiB Memory, 1 Tesla T4 GPU, 16 GiB GPU Memory), g4dn.8xlarge (32 vCPU, 128 GiB Memory, 1 Tesla T4 GPU, 16 GiB GPU Memory), p3.2xlarge (8 vCPU, 61 GiB Memory, 1 Tesla V100 GPU, 16 GiB GPU Memory), on the entire test dataset as shown in Table \ref{table:throughput-unstructured}.
The CNN model is the highest performing model on both CPU \& GPU instances. The next highest throughput models are the CuDNNLSTM and CNN-CuDNNLSTM models which are 2x-5x slower as they contain the LSTM layers. Note that unlike CuDNNLSTM layers, the regular LSTM layers can be utilized on the CPU, but are substantially slower on GPU.

CRF-based models (CNN-CRF, BiLSTM-CRF) suffer low throughput. Additionally, since the CRF layer relies on an RNN implementation, the models CRF-CNN, LSTM, BiLSTM-CRF, and CNN-LSTM have similar throughput. The CRF model with the lowest throughput is Ngram-CRF, likely due to its implementation using the sklearn-crfsuite package which supports CPU only.

For spaCy model, we configured the data pipeline to achieve maximum throughput. This data pipeline is different from our accuracy evaluation, and could result in slightly reduced accuracy. The spaCy model obtained higher throughput on the GPU rather than CPU, but is one of the slowest lowest performing models. Note that the SpaCy library uses the CuPy to execute the graph on GPU and does not take advantage of optimizations provided by Tensorflow, such as CuDNN.
\subsection{Column-wise Entity Prediction on Tabular Datasets}
In this section, we evaluate only structured datasets whose columns contain consistent entity formats. Our goal is to predict the entity type of each column. The columnar-level models along with the CNN model, the most optimal model in terms of accuracy and throughput, are evaluated.
\subsubsection{Accuracy Evaluation}
\begin{figure}[!htb]
\begin{center}
\centerline{\includegraphics[width=0.95\columnwidth]{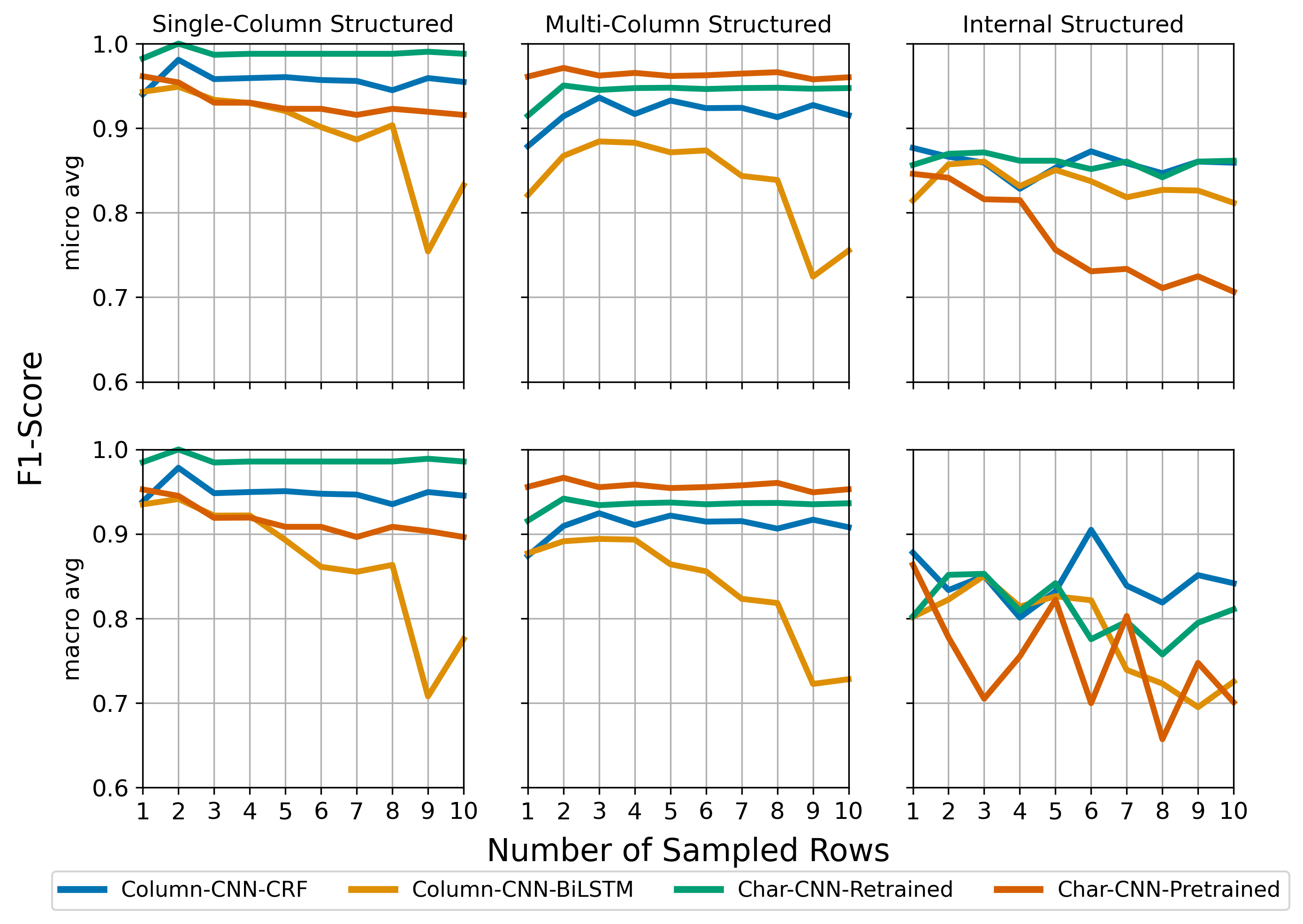}}
\caption{Accuracy (F1-score) for columnar-level models}
\label{fig:acc-structured}
\end{center}
\end{figure}
The accuracy results for column-wise entity prediction are given in Figure \ref{fig:acc-structured}.
Char-CNN-Retrained, Char-CNN-Pretrained and Column-CNN-CRF models obtained the best macro-average f1-score which depicts how the Char-CNN-Retrained and Char-CNN-Pretrained are context and task invariant after slight modifications to the data processing pipeline. Additionally, increasing the number of sampled rows does not appear to improve model accuracy. However, improvement via increased sampled rows may be dependent on the complexity of each entity as seen in the breakdown of accuracy results for individual sensitive entities given in the Appendix.
\subsubsection{Throughput Evaluation}
\begin{figure}[!htb]
\begin{center}
\centerline{\includegraphics[width=0.95\columnwidth]{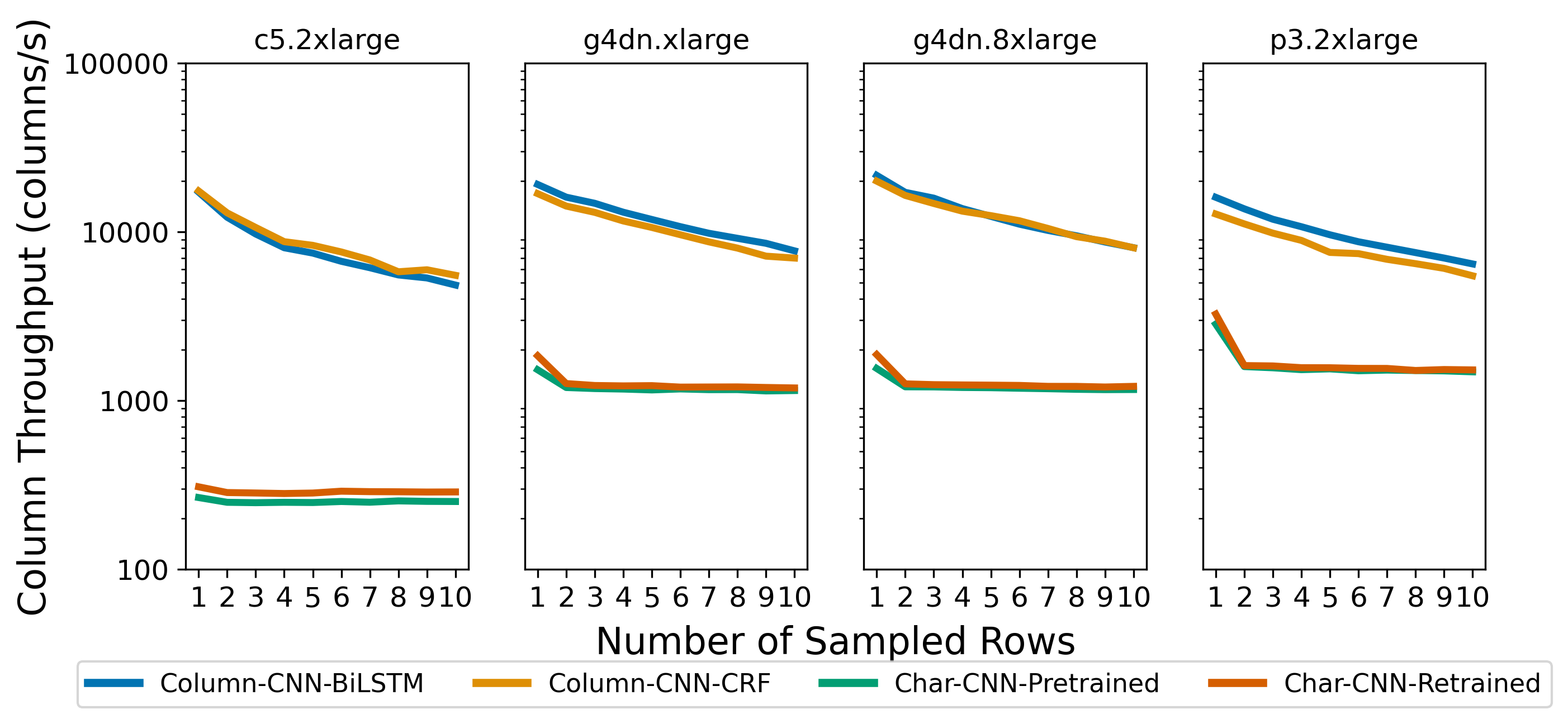}}
\caption{Throughput evaluation for columnar-level models.}
\label{fig:throughput-structured}
\end{center}
\end{figure}

Figure \ref{fig:throughput-structured} illustrates model throughput for the column-wise prediction task.
Despite similar accuracy performance with Column-CNN-CRF, the Char-CNN models had 3x-10x lower throughput. Throughput differences may be attributed to the larger parameter count of the Char-CNN models due to both CNN hyperparameter differences such as filter-size, num-filters, num-conv, and the output layer being significantly larger because of the character level prediction. Additionally, Char-CNN models did not use flattening mechanism which could also contribute to the throughput decreasing.
\section{Conclusions}
In this paper, we tackle the problems of identifying the sensitive information in different data formats for financial institutions. We optimize a set of neural network models to be deployed in the production environment(s). Our evaluation results on both synthetic data, real email and internal data shows that the CNN model is simple yet very effective with respect to accuracy and throughput, and thus the most suitable model for the production. We believe this work will shed some light on this challenging problem from which several lessons learned along with future directions are discussed as follows:
\begin{itemize}
\item \textit{Limitations of experimental data} The synthetic data is generated without context which may reduce the model performance on the real data with context, e.g., email datasets. Further directions may include the data generation with more realistic background which can be generated from some generative models such as GAN.
\item \textit{Data labeling} There exists some discrepancy in data labeling for the real email dataset, which might affect the overall results. However, it is worth noting that this is an intrinsic and common problem for this framework.
\item \textit{Overlaps of entities} There exist overlap values among several entities such as BAN, Phone-Number and SSN. Unless extra data (e.g., column names and statistical attributes of the value ranges) is provided, this problem appears inevitable in this framework.
\end{itemize}
\nonfrenchspacing
\bibliography{refs}
\bibliographystyle{aaai}
\clearpage
\appendix
\section{Appendix A: Examples of Entities}\label{app:entities}
\textbf{Sensitive entities}:\\
\textbf{Address (ADDRESS)}: 123 Fake st $\vert$ Houston, TX, 12345 $\vert$ 123 Fake st, Houston, TX, 12345 $\vert$ PO Box 123 Houston, TX\\
INVALID: Houston, TX | Texas | TX | Houston\\
\textbf{Bank Account Number (BAN)}: Numbers with 10-18 digits.\\
\textbf{Credit Card Number (CREDIT\_CARD)}: 4123-4567-8901-2345 $\vert$ 4123456789012345 $\vert$ 4123 4567 8901 2345 $\vert$ 4123.4567.8901.2345\\
\textbf{Datetime (DATETIME)}: July 2007 $\vert$ Tue, 03 Sep 2002 01:10:08 -0800 $\vert$ July 3 $\vert$ Tue, 03 Sep 2002 01:10:08 -0800 (PDT)\\
INVALID: July”, “2007 | Tuesday\\
\textbf{Email Address (EMAIL\_ADDRESS)}: fake-email@fake.com $\vert$ mailto:fake-email@fake.com $\vert$  mailto:fake-email@fake.com?subject=help $\vert$ fake-email@localhost\\
\textbf{Hash or Encryption Key (HASH\_OR\_KEY)}: 43AB87CD0872347F89E87E87F\\
\textbf{IPv4 (IPV4)}: 123.45.67.89 | 12.34.123.123\\
\textbf{IPv6 (IPV6)}: 1234:f123:4567:89d:bcdf:c123:bbb0:123d $\vert$ 12b3:4d5f:123a:b4c5::/67\\
\textbf{MAC Address (MAC\_ADDRESS)}: 12-3A-BC-DE-FA-B0 $\vert$ 12:3A:BC:DE:FA:B0 $\vert$ 123ABCDEFAB0 $\vert$ 123A.BCDE.FAB0\\
\textbf{Person’s Name (PERSON)}: Jane $\vert$ John $\vert$ Smith $\vert$ Jane Smith $\vert$ John X Smith $\vert$ Mr. John Smith $\vert$ Prof. Smith $\vert$ Gov. Smith, Jane $\vert$ John’s $\vert$, Dr. Jane’s\\
\textbf{Phone Number (PHONE\_NUMBER)}: 123-456-7890 x12345 $\vert$ +1 1234567890 $\vert$ 001-123-456-7890 $\vert$ +1 (123) 456-7890 $\vert$ 456-7890 $\vert$ 123-456-7890 12345\# $\vert$ x12345, (12345)\\
\textbf{Social Security Number (SSN)}: 123-45-6789 $\vert$ 123.45.6789 $\vert$ 123456789 $\vert$ 123 45 6789\\
\textbf{URL (URL)}: https://google.com/examples?test=1 $\vert$ www.google.com/examples?test=1 $\vert$ google.com/examples?test=1 \\
Valid schemes: ftp:// $\vert$ http:// $\vert$ https:// $\vert$ s3:// $\vert$ rpc:// \\
\textbf{UUID (UUID)}: 00000000-0000-4000-a000-000000000000\\\\
\textbf{Non-Sensitive Entities}:\\
\textbf{Background (BACKGROUND)}: Any text which does not fall into the other categories\\
\textbf{Float (FLOAT)}: 1.0 $\vert$ 1. $\vert$ .1 $\vert$ (3.1) $\vert$ [3.1] $\vert$ 3.1) $\vert$ 1./3.0\\
\textbf{Integer (INTEGER)}: 3 $\vert$ (3) $\vert$ [3] $\vert$ 3) $\vert$ 3! $\vert$ 1/3\\
\textbf{Ordinal Value (ORDINAL)}: 4th $\vert$ Second $\vert$ 1) $\vert$ 2. $\vert$ v3.1 $\vert$ v3.1.1, 3.1.1 6.x\\
\textbf{Quantity (QUANTITY)}: \$432.23 $\vert$ 4bits $\vert$ 4-bits\\
INVALID: \$ 432.23 $\vert$ 4 bits
\section{Appendix B: Regex Patterns}
A detailed example of the regex pattern:\\
{\scriptsize
\begin{verbatim}
regex_encapsulators = {
   'start': r'(?<![\w.\$\%\-])',
   'end': r'(?:(?=(\b|[ ]))|(?=[^\w\%\$]([^\w]|$))|$)',
}
\end{verbatim}
}
\lstset{breaklines=true}
{\scriptsize
\begin{lstlisting}
ADDRESS
'\d{1,3}.?\d{0,3}\s[a-zA-Z]{2,30}\s[a-zA-Z]{2,15}.?',
'((?:(?:\d+(?:\x20+\w+?\.?)+?(?:\x20+(?:STREET|ST|DRIVE|DR|AVENUE|AVE|ROAD|RD|LOOP|COURT|CT|CIRCLE|LANE|LN|BOULEVARD|BLVD)\.?))|(?:(?:P\.\x20?O\.|P\x20?O)\x20*Box\x20+\d+)|(?:General\x20+Delivery)|(?:C[\\\/]O\x20+(?:\w+\x20*)+))\,?\x20*?(?:(?:(?:APT|BLDG|DEPT|FL|HNGR|LOT|PIER|RM|S(?:LIP|PC|T(?:E|OP))|TRLR|UNIT|\x23)\.?\x20*?(?:[a-zA-Z0-9\-]+))|(?:BSMT|FRNT|LBBY|LOWR|OFC|PH|REAR|SIDE|UPPR))?)(\,?\s+(?:(?:\d+(?:\x20+\w+\.?)+?(?:\x20+(?:STREET|ST|DRIVE|DR|AVENUE|AVE|ROAD|RD|LOOP|COURT|CT|CIRCLE|LANE|LN|BOULEVARD|BLVD)\.?)?)|(?:(?:P\.\x20?O\.|P\x20?O)\x20*Box\x20+\d+)|(?:General\x20+Delivery)|(?:C[\\\/]O\x20+(?:\w+\x20*)+))\,?\x20*?(?:(?:(?:APT|BLDG|DEPT|FL|HNGR|LOT|PIER|RM|S(?:LIP|PC|T(?:E|OP))|TRLR|UNIT|\x23)\.?\x20*(?:[a-zA-Z0-9\-]+))|(?:BSMT|FRNT|LBBY|LOWR|OFC|PH|REAR|SIDE|UPPR))?)?(\,?\s+(?:[A-Za-z]+\x20*?)+)?(\,?\s+(?:[Aa](?:la(?:(?:bam|sk)a)|merican [Ss]amoa|r(?:izona|kansas))|(?:^(?![Bb]aja )[Cc]alifornia)|[Cc]o(?:lorado|nnecticut)|[Dd](?:elaware|istrict of [Cc]olumbia)|[Ff]lorida|[Gg](?:eorgia|uam)|[Hh]awaii|[Ii](?:daho|llinois|ndiana|owa)|[Kk](?:ansas|entucky)|[Ll]ouisiana|[Mm](?:a(?:ine|ryland|ssachusetts)|i(?:chigan|nnesota|ss(?:(?:issipp|our)i))|ontana)|[Nn](?:e(?:braska|vada|w (?:[Hh]ampshire|[Jj]ersey|[Mm]exico|[Yy]ork))|orth (?:(?:[Cc]arolin|[Dd]akot)a))|[Oo](?:hio|klahoma|regon)|[Pp](?:ennsylvania|uerto [Rr]ico)|[Rr]hode [Ii]sland|[Ss]outh (?:(?:[Cc]arolin|[Dd]akot)a)|[Tt](?:ennessee|exas)|[Uu]tah|[Vv](?:ermont|irgin(?:ia| [Ii]sland(s?)))|[Ww](?:ashington|est [Vv]irginia|isconsin|yoming)|A[KLRSZ]|C[AOT]|D[CE]|FL|G[AU]|HI|I[ADLN]|K[SY]|LA|M[ADEINOST]|N[CDEHJMVY]|O[HKR]|P[AR]|RI|S[CD]|T[NX]|UT|V[AIT]|W[AIVY]))?(\s+\d{4,5}(?:-\d{4})?)?',
'((?:[A-Za-z]+\x20*){1,3}?)(\,?\s+(?:[Aa](?:la(?:(?:bam|sk)a)|merican [Ss]amoa|r(?:izona|kansas))|(?:^(?![Bb]aja )[Cc]alifornia)|[Cc]o(?:lorado|nnecticut)|[Dd](?:elaware|istrict of [Cc]olumbia)|[Ff]lorida|[Gg](?:eorgia|uam)|[Hh]awaii|[Ii](?:daho|llinois|ndiana|owa)|[Kk](?:ansas|entucky)|[Ll]ouisiana|[Mm](?:a(?:ine|ryland|ssachusetts)|i(?:chigan|nnesota|ss(?:(?:issipp|our)i))|ontana)|[Nn](?:e(?:braska|vada|w (?:[Hh]ampshire|[Jj]ersey|[Mm]exico|[Yy]ork))|orth (?:(?:[Cc]arolin|[Dd]akot)a))|[Oo](?:hio|klahoma|regon)|[Pp](?:ennsylvania|uerto [Rr]ico)|[Rr]hode [Ii]sland|[Ss]outh (?:(?:[Cc]arolin|[Dd]akot)a)|[Tt](?:ennessee|exas)|[Uu]tah|[Vv](?:ermont|irgin(?:ia| [Ii]sland(s?)))|[Ww](?:ashington|est [Vv]irginia|isconsin|yoming)|A[KLRSZ]|C[AOT]|D[CE]|FL|G[AU]|HI|I[ADLN]|K[SY]|LA|M[ADEINOST]|N[CDEHJMVY]|O[HKR]|P[AR]|RI|S[CD]|T[NX]|UT|V[AIT]|W[AIVY]))(\s+\d{4,5}(?:-\d{4})?)(?:(?=(\b|[ ]))|(?=[^\w\%\$]([^\w]|$))|$)'
BAN
'\d{10,18}'
CREDIT_CARD
'(?:3[74]\d{13}|[456]\d{15})'
'[3456]\d{2,3}[^a-zA-Z0-9]\d{4}[^a-zA-Z0-9]\d{4}[^a-zA-Z0-9]\d{4}'
DATETIME
'([0]\d|[1][0-2])\/([0-2]\d|[3][0-1])\/([2][01]|[1][6-9])\d{2}(\s([0-1]\d|[2][0-3])(\:[0-5]\d){1,2})?'
'(?:0?[0-9]|1[012]|2[0123])((?:\:[0-5]\d){1,2}(?:\x20?[ap]m)?|\x20?[ap]m)'
'(?:Jan(uary)?|Feb(ruary)?|Mar(ch)?|Apr(il)?|May|Jun(e)?|Jul(y)?|Aug(ust)?|Sep(tember)?|Oct(ober)?|Nov(ember)?|Dec(ember)?)\s+(?:(\d{1,2}\,?\s+\d{4})|(\d{1,4}))' # July 23, 2007\July 2007\ July 23
'(\d{1,2}\s+)(?:Jan(uary)?|Feb(ruary)?|Mar(ch)?|Apr(il)?|May|Jun(e)?|Jul(y)?|Aug(ust)?|Sep(tember)?|Oct(ober)?|Nov(ember)?|Dec(ember)?)(\d{1,4})?' # 23 July\ 23 July 2007
'((?:mon|tue(?:s)?|wed(?:nes)?|thur(?:s)?|fri|sat(?:ur)?|sun)(?:day)?\,?\s+)?(\d{1,2}\s+)?(?:Jan(uary)?|Feb(ruary)?|Mar(ch)?|Apr(il)?|May|Jun(e)?|Jul(y)?|Aug(ust)?|Sep(tember)?|Oct(ober)?|Nov(ember)?|Dec(ember)?)\s+(?:(\d{1,2}\,?\s+\d{4})|(\d{1,4}))(\s+\d{2}\:\d{2}\:\d{2}(\s+[+-]?\d{4})?(\s+\([a-zA-Z]{3}\))?)' # 03 Sep 2002 01:10:08 -0800 (DFT)\Sep 03 2002 01:10:08 -0800 (DFT)\Sep 03 2002 01:10:08
EMAIL_ADDRESS
'(?<=mailto\:)?(?:[a-zA-Z0-9!#\$\%&\'*+/=?^_`{|}~-]+(?:\.+[a-zA-Z0-9!#\$\%&\'*+/=?^_`{|}~-]+)*|\"(?:[\x01-\x08\x0b\x0c\x0e-\x1f\x21\x23-\x5b\x5d-\x7f]|\\[\x01-\x09\x0b\x0c\x0e-\x7f])*\")@(?:(?:[a-zA-Z0-9](?:[a-zA-Z0-9-]*[a-zA-Z0-9])?\.)+[a-zA-Z0-9](?:[a-zA-Z0-9-]*[a-zA-Z0-9])?|\\[(?:(?:25[0-5]|2[0-4][0-9]|[01]?[0-9][0-9]?)\.){3}(?:25[0-5]|2[0-4][0-9]|[01]?[0-9][0-9]?|[a-zA-Z0-9-]*[a-zA-Z0-9]:(?:[\x01-\x08\x0b\x0c\x0e-\x1f\x21-\x5a\x53-\x7f]|\\[\x01-\x09\x0b\x0c\x0e-\x7f])+)\]|localhost)'
FLOAT
[+-]?(?:(?:[0-9]+([\.][0-9]*(e[\+-]?[0-9]+)?))|(?:[\.][0-9]+))(?!([\.-_]\d|\d\.))
HASH_OR_KEY
'([a-zA-Z\d\-\=\\\/\+]{16,})' # 16 alpha numeric including some special chars
INTEGER
'[-+]?\d+(?![\.-_]\d)'
IPV4
r'(((25[0-5]|(2[0-4]|1[0-9]|[1-9]|)[0-9])(\.(?!\$)|(?=\b))){4}|localhost)' # https://stackoverflow.com/questions/5284147/validating-ipv4-addresses-with-regexp
IPV6
'(([0-9a-fA-F]{1,4}:){7,7}[0-9a-fA-F]{1,4}|([0-9a-fA-F]{1,4}:){1,7}:|([0-9a-fA-F]{1,4}:){1,6}:[0-9a-fA-F]{1,4}|([0-9a-fA-F]{1,4}:){1,5}(:[0-9a-fA-F]{1,4}){1,2}|([0-9a-fA-F]{1,4}:){1,4}(:[0-9a-fA-F]{1,4}){1,3}|([0-9a-fA-F]{1,4}:){1,3}(:[0-9a-fA-F]{1,4}){1,4}|([0-9a-fA-F]{1,4}:){1,2}(:[0-9a-fA-F]{1,4}){1,5}|[0-9a-fA-F]{1,4}:((:[0-9a-fA-F]{1,4}){1,6})|:((:[0-9a-fA-F]{1,4}){1,7}|:)|fe80:(:[0-9a-fA-F]{0,4}){0,4}\%[0-9a-zA-Z]{1,}|::(ffff(:0{1,4}){0,1}:){0,1}((25[0-5]|(2[0-4]|1{0,1}[0-9]){0,1}[0-9])\.){3,3}(25[0-5]|(2[0-4]|1{0,1}[0-9]){0,1}[0-9])|([0-9a-fA-F]{1,4}:){1,4}:((25[0-5]|(2[0-4]|1{0,1}[0-9]){0,1}[0-9])\.){3,3}(25[0-5]|(2[0-4]|1{0,1}[0-9]){0,1}[0-9]))(\%.+)?s*(\/([0-9]|[1-9][0-9]|1[0-1][0-9]|12[0-8]))?' # https://stackoverflow.com/questions/53497/regular-expression-that-matches-valid-ipv6-addresses
MAC_ADDRESS
'([0-9a-fA-F][0-9a-fA-F][:.-]?){5}([0-9a-fA-F][0-9a-fA-F])'
PERSON
No regex for person was created.
PHONE_NUMBER
'((?:\+?(\d{1,3}))?[-. ]*\\(?(\d{3})\)?[-. ]?\\(?(\d{3})\)?[-. ]*(\d{4})(?:[\x20\,#]*(?:x|Ext\.?|#)\d+)?|(?:x|Ext\.?|#)\d+)' # https://stackoverflow.com/questions/16699007/regular-expression-to-match-standard-10-digit-phone-number
QUANTITY
'(?<![0-9][\:\-])[+-]?([0-9]+([.][0-9]*)?|[.][0-9]+)[a-zA-Z\%-]*[a-zA-Z\%]+(?!(\+[0-9]+))(?!([\-\@][0-9a-zA-Z]+[\-\.]))' # int or float followed by text
'\$[+-]?(([0-9]\,?)+([.][0-9]*)?|[.][0-9]+)'  # preceding \$ int or float
ORDINAL
r'(?<!\d\.)(v(?:\d+\.)*\d+[a-zA-Z]*|(?:\d+\.){2}\d+|(?:\d+\.)+(?:\d+(?:[a-zA-Z]+)))(?!\.+[a-zA-Z0-9]+|[0-9]+(\.+|[a-zA-Z]))(?:\-\d+)?'  # versioning: 1.1.1, v1.1.1
r'(?:\d*(?:1st|2nd|3rd)|\d*(?:1.|[04-9])th)'  # ordinal numbers 1st, 22nd, etc.
r'(?<!\\()(?:\#\d{1,3}|\d{1,3}[\.\)\:])(?=\W|\$)'  # bulleted numbers e.g. 1), 3.
r'(?<!\@)((?:one|t(?:w(?:o|e(?:lve|nty))|h(?:ree|irt(?:een|y))|en)|f(?:our(?:teen|ty)?|ive|if(?:teen|ty))|s(?:ix|even)(?:teen|ty)?|e(?:ight(?:een|y)?|leven)|nine(?:teen|ty)?|hundred|thousand|(?:m|b|tr|quadr)illion)[\W-]?)*((?:(?:first|second|third)|(?:f(?:ou?r|if)(?:teen|tie)?|s(?:ix|even)(?:teen|tie)?|e(?:igh(?:teen|tie)?|leven)|t(?:we(?:lf|ntie)|en)|thirt(?:een|ie)|nin(?:e(?:teen|tie))?|hundre|thousand|(?:m|b|tr|quadr)illion)th))'  # ordinal words
SSN
'\d{9}'
'\d{3}[^a-zA-Z0-9]\d{2}[^a-zA-Z0-9]\d{4}'
URL
'(?:https?:\/\/(www\.)?|(rpc|s3|ftp):\/\/)?[a-zA-Z0-9]+(?:[\-\.]{1}[a-zA-Z0-9]+)*\.[a-zA-Z]{2,5}(:[0-9]{1,5})?(?:\/(?:[\S])*)?[a-zA-Z0-9\/]'
UUID
'[a-f0-9]{8}-?[a-f0-9]{4}-?4[a-f0-9]{3}-?[89ab][a-f0-9]{3}-?[a-f0-9]{12}' # https://stackoverflow.com/questions/11384589/what-is-the-correct-regex-for-matching-values-generated-by-uuid-uuid4-hex
\end{lstlisting}
}
\section{Appendix C: Detailed Parameters for Unstructured Text Generation}
General tunable parameters to generate unstructured text are given as:\\\\
\textit{num-samples}:  number of text samples generated\\
\textit{bg-word-count-min}: minimum background words\\
\textit{bg-word-count-max}: maximum background words\\
\textit{prob-of-npi}: probability the word is non-background\\
\textit{prob-two-npi-together}: probability two non-background occur consecutively\\
\textit{prob-json}: probability of format json being selected\\
\textit{prob-structured}: probability of format structured being selected\\
\textit{prob-sentence}: automatically determined by the subtraction of the prob-json and prob-structured from 1.0

For each background word, the brown corpus distribution was used for randomly select a list and subsequently a word from that list. In addition, capitalization was applied to each word with distribution: 'lower'=0.59, 'upper'=0.20, 'title'=0.20, 'random'=0.01. Finally, background words were connected together by a random delimiter ('', '\_', '-', '.') with a distribution of 1-4 connected words equaling (0.9, 0.065, 0.025, 0.01) respectively.
\begin{table*}[thb]
\scriptsize
\centering\renewcommand\cellalign{lc}
\setcellgapes{3pt}\makegapedcells
\begin{tabular}{|l|l|l|l|l|l|l|l|}
\hline
\textbf{Data}  & \textbf{num-samples} & \textbf{bg-word-count-min} & \textbf{bg-word-count-max} & \textbf{prob-of-npi} & \textbf{prob-two-npi-together} & \textbf{prob-json} & \textbf{prob}\\
\hline
\multirow{6}{*}{Train} & 35000 & 10 & 40 & 0.25 & 0.2 & 0.05 & 0.25\\
& 3000 & 2 & 50 & 0.4 & 0.1 & 0.1 & 0.5\\
& 3000 & 2 & 50 & 0.4 & 0.1 & 0 & 0\\
& 3000 & 10 & 30 & 0.1 & 0.4 & 0.05 & 0.3\\
& 3000 & 1 & 3 & 0.5 & 0.1 & 0 & 0.5\\
& 3000 & 10 & 30 & 0.2 & 0.2 & 0 & 0\\
\hline
\multirow{6}{*}{Test} & 350 & 10 & 40 & 0.25 & 0.2 & 0.05 & 0.25\\
& 300 & 2 & 50 & 0.4 & 0.1 & 0.1 & 0.5\\
& 300 & 2 & 50 & 0.4 & 0.1 & 0 & 0\\
& 1200 & 10 & 30 & 0.1 & 0.4 & 0.05 & 0.3\\
& 3000 & 1 & 3 & 0.5 & 0.1 & 0 & 0.5\\
& 600 & 10 & 30 & 0.2 & 0.2 & 0 & 0\\
\hline
\end{tabular}
\caption{Six variations of tunable parameters used to generate approximately 125k entities of unstructured text training data, and 20k entities of unstructured text for testing data}
\label{table:traindata-unstructure}
\end{table*}
\section{Appendix D: Accuracy Results for Entity Detection Task for All Entities}\label{app:acc-all-entity}
\begin{table*}[thb]
\scriptsize
\centering\renewcommand\cellalign{lc}
\setcellgapes{3pt}\makegapedcells
\begin{tabular}{|l|l|l|l|l|l|}
\hline
\textbf{Model / Datasets}   & \makecell{\textbf{Multi-column,} \\ \textbf{Structured Data}}   & \makecell{\textbf{Single-column,} \\ \textbf{Structured Data}}   & \makecell{\textbf{Unstructured Text} \\ \textbf{}}   & \makecell{\textbf{Public Emails} \\ \textbf{}}   & \makecell{\textbf{Internal,} \\ \textbf{Structured  Data}}\\
\hline
Char CNN & \makecell{(0.99, 0.99, 0.99) \\ (0.97, 0.97, 0.97)} & \makecell{(0.98, 0.98, 0.98) \\ (0.96, 0.95, 0.96)} & \makecell{(0.97, 0.96, 0.97) \\ (0.95, 0.93, 0.94)} & \makecell{(0.65, 0.76, 0.70) \\ (0.49, 0.67, 0.54)} & \makecell{(0.83, 0.83, 0.83) \\ (0.75, 0.82, 0.74)} \\
\hline
Char CuDNNLSTM  & \makecell{(0.94, 0.92, 0.93) \\ (0.88, 0.86, 0.86)} & \makecell{(0.94, 0.92, 0.93) \\ (0.86, 0.81, 0.83)} & \makecell{(0.83, 0.79, 0.81) \\ (0.80, 0.69, 0.72)} & \makecell{(0.53, 0.59, 0.56) \\ (0.42, 0.54, 0.44)} & \makecell{(0.56, 0.61, 0.58) \\ (0.57, 0.61, 0.54)} \\
\hline
Char CNN + CuDNNLSTM  & \makecell{(0.99, 0.98, 0.98) \\ (0.96, 0.95, 0.96)} & \makecell{(0.99, 0.98, 0.99) \\ (0.95, 0.95, 0.95)} & \makecell{(0.97, 0.95, 0.96) \\ (0.94, 0.91, 0.92)} & \makecell{(0.66, 0.73, 0.69) \\ (0.52, 0.63, 0.53)} & \makecell{(0.85, 0.85, 0.85) \\ (0.78, 0.83, 0.76)} \\
\hline
Word SpaCy  & \makecell{(0.86, 0.84, 0.85) \\ (0.82, 0.74, 0.77)} & \makecell{(0.96, 0.96, 0.96) \\ (0.93, 0.90, 0.90)} & \makecell{(0.75, 0.76, 0.76) \\ (0.74, 0.70, 0.71)} & \makecell{(0.62, 0.62, 0.62) \\ (0.52, 0.60, 0.50)} & \makecell{(0.48, 0.66, 0.56) \\ (0.53, 0.67, 0.48)} \\
\hline
Char Ngram + CRF & \makecell{(0.96, 0.94, 0.95) \\ (0.90, 0.88, 0.88)} & \makecell{(0.96, 0.96, 0.96) \\ (0.90, 0.86, 0.87)} & \makecell{(0.91, 0.88, 0.89) \\ (0.83, 0.80, 0.81)} & \makecell{(0.60, 0.69, 0.64) \\ (0.44, 0.61, 0.48)} & \makecell{(0.79, 0.79, 0.79) \\ (0.72, 0.75, 0.70)} \\
\hline
Char CNN + CRF & \makecell{(0.99, 0.99, 0.99) \\ (0.97, 0.98, 0.97)} & \makecell{(0.98, 0.98, 0.98) \\ (0.95, 0.94, 0.94)} & \makecell{(0.98, 0.97, 0.97) \\ (0.96, 0.95, 0.95)} & \makecell{(0.69, 0.79, 0.73) \\ (0.54, 0.70, 0.58)} & \makecell{(0.85, 0.87, 0.86) \\ (0.74, 0.83, 0.75)} \\
\hline
Char BiLSTM + CRF & \makecell{(0.99, 0.98, 0.99) \\ (0.97, 0.96, 0.97)} & \makecell{(0.99, 0.98, 0.98) \\ (0.95, 0.93, 0.94)} &  \makecell{(0.97, 0.95, 0.96) \\ (0.95, 0.93, 0.93)} & \makecell{(0.69, 0.77, 0.73) \\ (0.54, 0.70, 0.58)} & \makecell{(0.83, 0.83, 0.83) \\ (0.77, 0.81, 0.73)} \\
\hline
Regex &  \makecell{(0.68, 0.73, 0.70) \\ (0.71, 0.64, 0.61)} & \makecell{(0.85, 0.80, 0.83) \\ (0.73, 0.59, 0.59)} & \makecell{(0.63, 0.67, 0.65) \\ (0.69, 0.62, 0.60)} & \makecell{(0.63, 0.63, 0.63) \\ (0.55, 0.67, 0.56)} & \makecell{(0.40, 0.61, 0.48) \\ (0.64, 0.60, 0.55)}\\
\hline
\end{tabular}
\caption{Evaluation results for five parts of the test set, with all entities. For each cell, the first line shows the micro average results and the second line shows the micro average results. Each line represents precision, recall and F1-score.}
\label{table:accuracy-all}
\end{table*}
This section provides the results for prediction of the models on all entities including the non-sensitives: INTEGER, FLOAT, QUANTITY, ORDINAL. Table \ref{table:accuracy-all} shows the accuracy results on the test set.
It can be observed that most results are almost the same as the accuracy for sensitive entities except for real email data where we see at least 2\% decrease in both micro and macro average of the F1-score for all models. In fact, the accuracy for non-sensitive entities are not high compared to other entities as they share some overlap regions. For example, ordinal list data (1., 2., etc) can be confused with float numbers, or quantity data (3 feet, 3.1 meters, etc) can be confused with float or integer as the labelers may rely on the context when labeling. Thus, omitting these non-sensitive entities in effect helps increase the overall accuracy.
\clearpage
\section{Appendix E: Accuracy Results for Entity Detection Task at the Entity Level}\label{app:acc-entity}
\begin{figure*}[!thb]
\begin{center}
\centerline{\includegraphics[width=1.5\columnwidth]{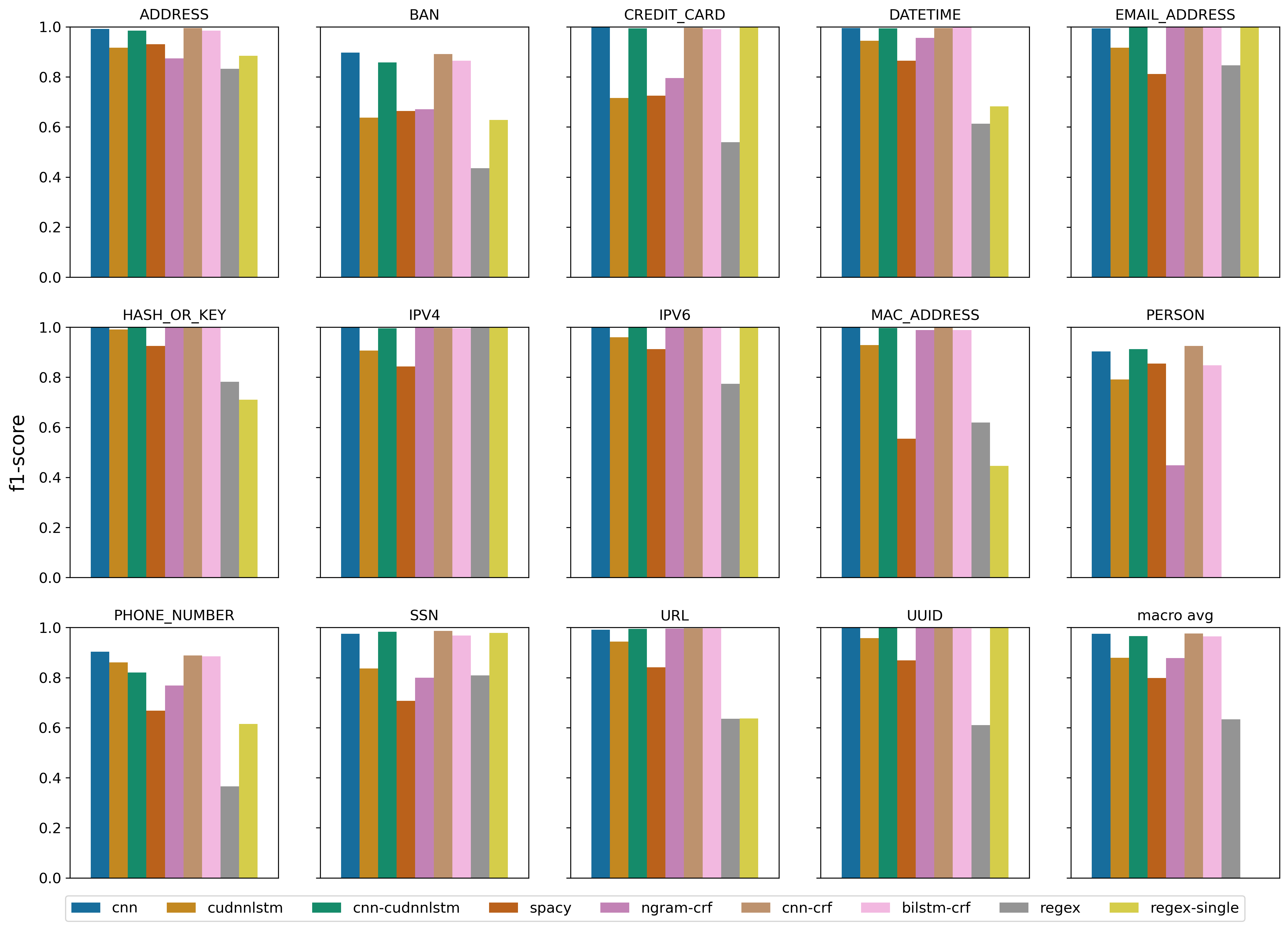}}
\caption{Accuracy breakdown at entity level for multi-column structured data.}
\label{fig:acc-entity-random-schema}
\end{center}
\end{figure*}
\begin{figure*}[!thb]
\begin{center}
\centerline{\includegraphics[width=1.5\columnwidth]{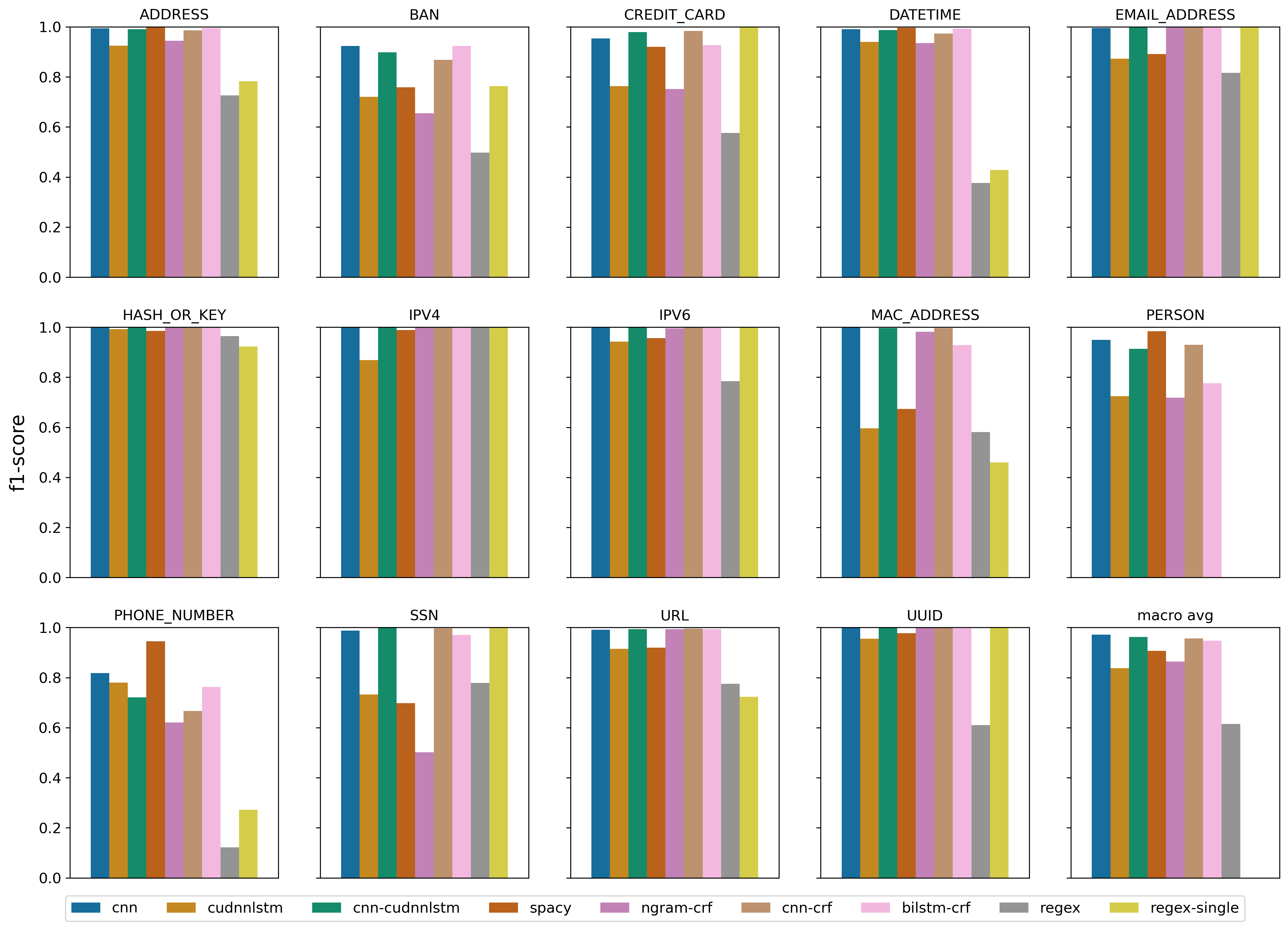}}
\caption{Accuracy breakdown at entity level for single-column structured data.}
\label{fig:acc-entity-one-column}
\end{center}
\end{figure*}
\begin{figure*}[!htb]
\begin{center}
\centerline{\includegraphics[width=1.5\columnwidth]{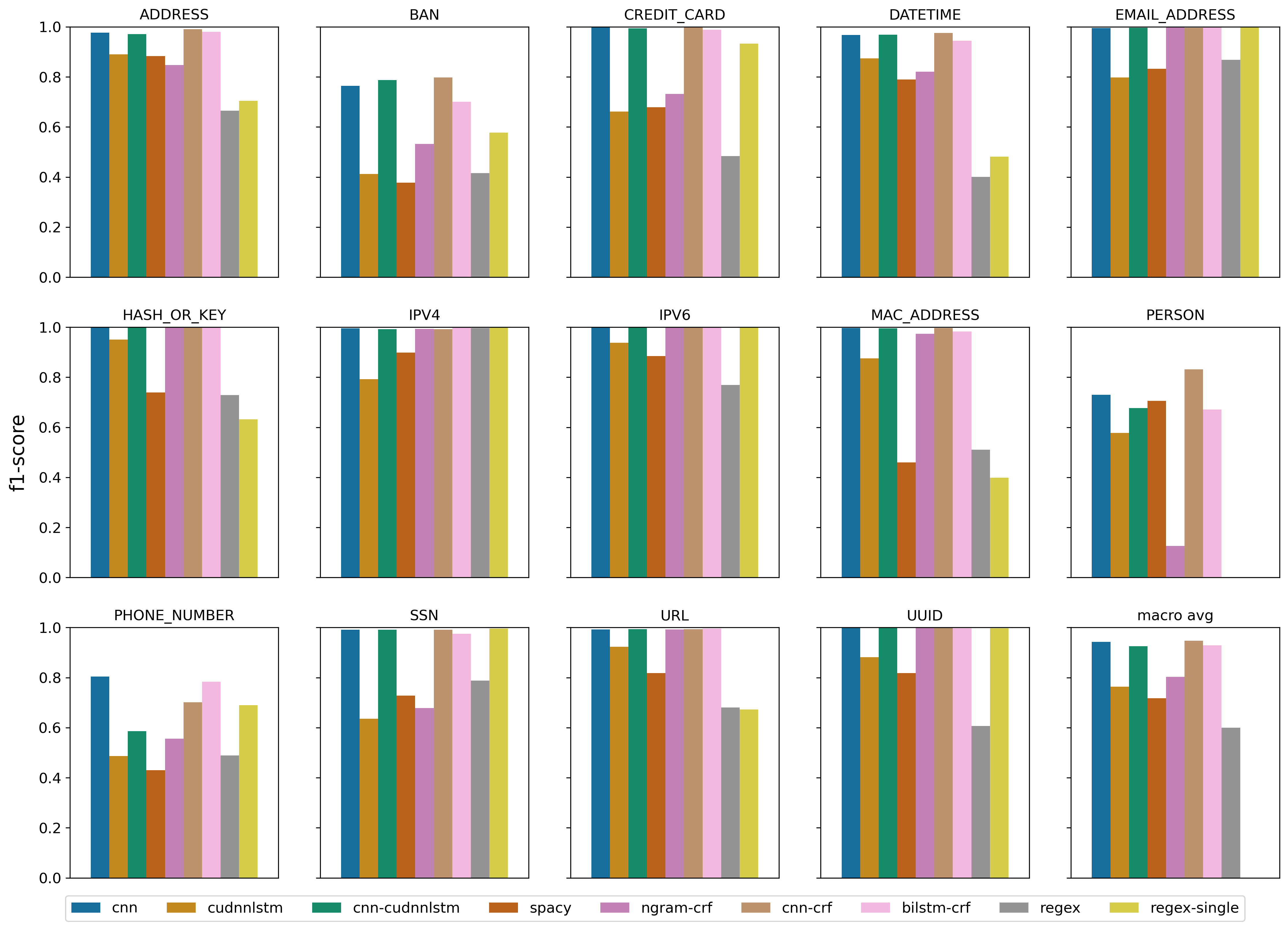}}
\caption{Accuracy breakdown at entity level for unstructured text.}
\label{fig:acc-entity-random-text}
\end{center}
\end{figure*}
\begin{figure*}[!htb]
\begin{center}
\centerline{\includegraphics[width=1.5\columnwidth]{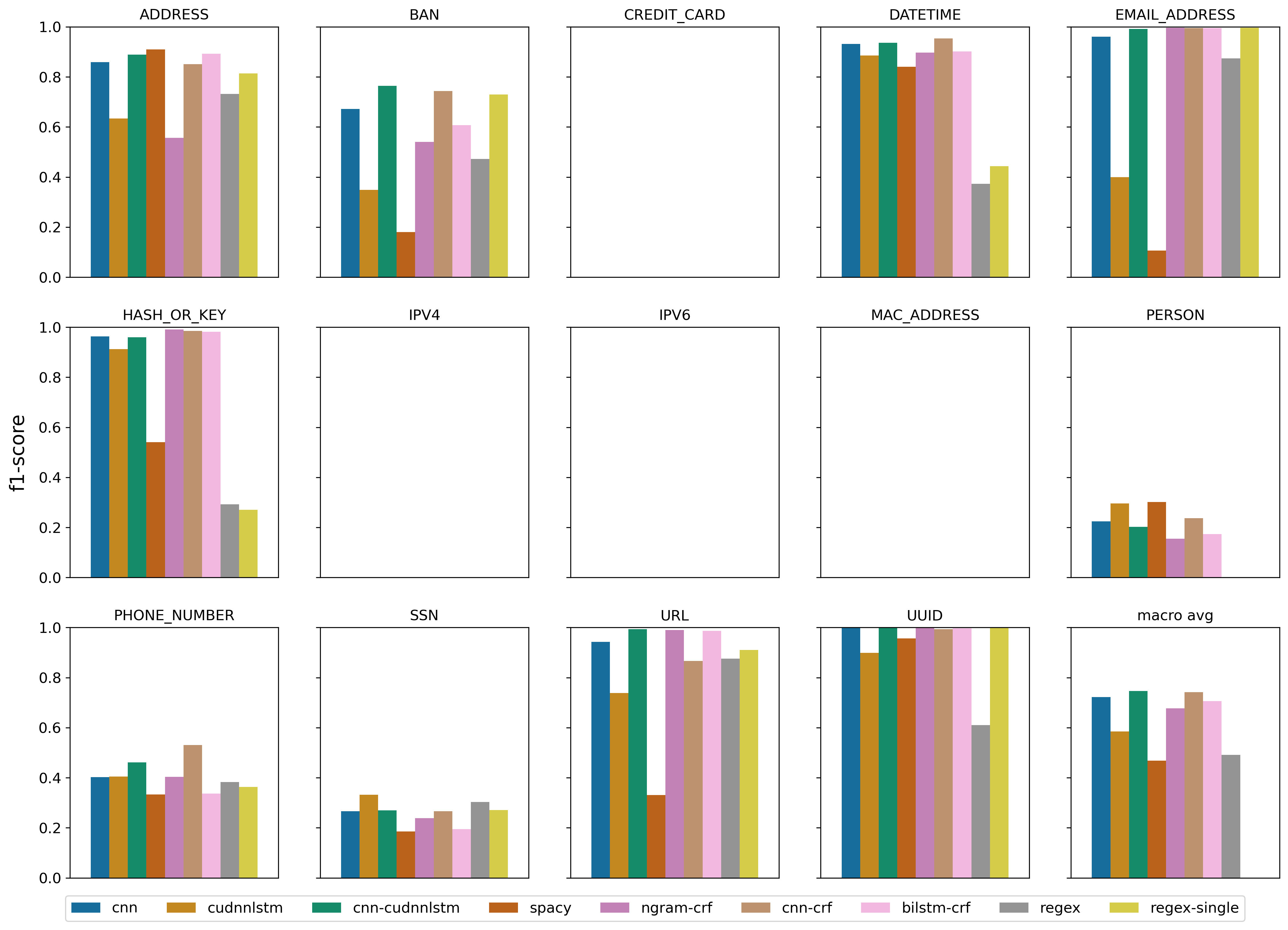}}
\caption{Accuracy breakdown at entity level for the internal structured data. Empty subplots indicate the non-existence of the sensitive entities.}
\label{fig:acc-entity-internal-data}
\end{center}
\end{figure*}
\begin{figure*}[!htb]
\begin{center}
\centerline{\includegraphics[width=1.5\columnwidth]{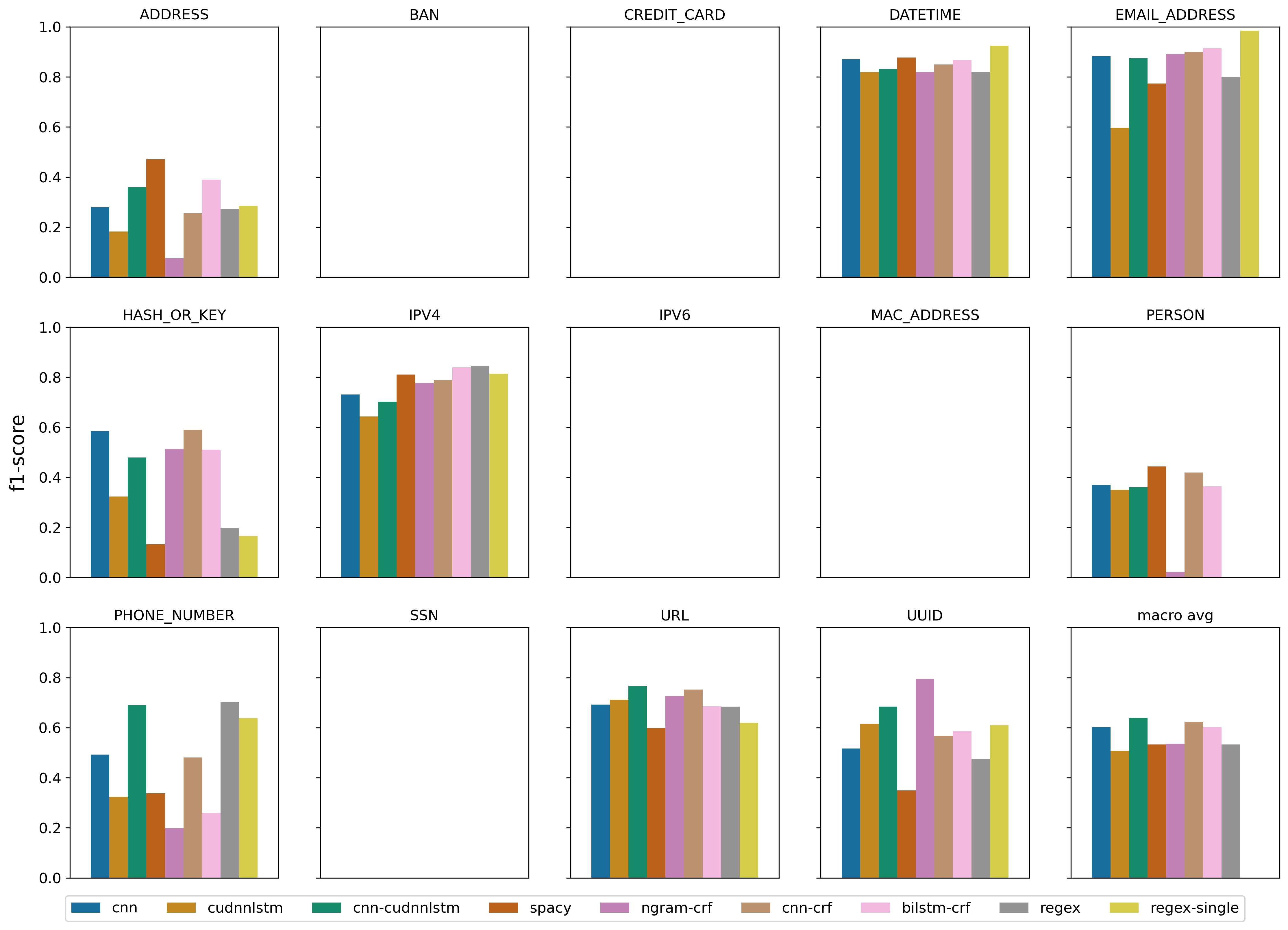}}
\caption{Accuracy breakdown at entity level for public emails. Empty subplots indicate the non-existence of the sensitive entities.}
\label{fig:acc-entity-email}
\end{center}
\end{figure*}
To further analyze the performance of the models on the real data, a breakdown of accuracy results for individual sensitive entities in five datasets are provided in Figure \ref{fig:acc-entity-random-schema}, Figure \ref{fig:acc-entity-one-column}, Figure \ref{fig:acc-entity-random-text}, Figure \ref{fig:acc-entity-internal-data} and Figure \ref{fig:acc-entity-email}, respectively.
One can observe from Figure \ref{fig:acc-entity-internal-data} that in general, CNN, CNN-CuDNNLSTM, CNN-CRF, and BiLSTM-CRF are the top models in which CNN-CuDNNLSTM and CNN-CRF slightly outperforms the rest, as stated above and also shown by the last subplot with the macro average F1-score. Across the entities, Datetime, Email-Address, Hash-Or-Key, URL, UUID receive much higher accuracy from most models, especially the top models, as those entities contain some special markers such as @ for emails or \textbackslash \textbackslash for url, and special patterns such as hexadecimal characters for uuid or colon separated datetimes. In contrast, BAN, Phone-Number, SSN obtain lower accuracy due to the inevitable overlap values of these entities: 10-digits phone numbers without separators (e.g., '()', '-') can be confused by a BAN (with 10, 11, 14 or 18 digits); 9-digits ssn without separators (e.g., '-') can be confused by a phone numbers without separators. It is worth noting here that these overlap regions appear unavoidable unless extra meta-data is given. Similarly, there exists some overlap between person names, addresses and background (person name can be a noun/background, and a city can be named after a person) that degrades the accuracy for these entities. As compared to the regex models, the top models obtain as good accuracy as the individual regex on most entities and even outperform some regex for Datetime (in the internal data) and Hash-Or-Key whose numerous formats cannot be fully captured using regex. As mentioned above, for the more fair comparison, the combined regex model distributes the entity prediction equally for each character. This combined regex model is obviously much worse than the individual regex models, and therefore is inferior to the top models.
The accuracy on the real emails, given in Figure \ref{fig:acc-entity-email}, is generally lower than that of the internal data, especially on Address, Hash-Or-Keys, URL and UUID as: Addresses are spread out to multiple lines; Hash-Or-Keys in the real text have wider range of lengths and special characters; UUID and URL may be included inside an email address. The accuracy for Phone-Number is better than in the internal data as there is no conflicting BAN or SSN in this data, but still low as there exists some extension numbers without the attached characters that may be confused with other numeric entities.
\clearpage
\section{Appendix F: Accuracy Results for Column-wise Predictions at the Entity Level}\label{app:acc-structured-entity}
\begin{figure*}[!htb]
\begin{center}
\centerline{\includegraphics[width=1.5\columnwidth]{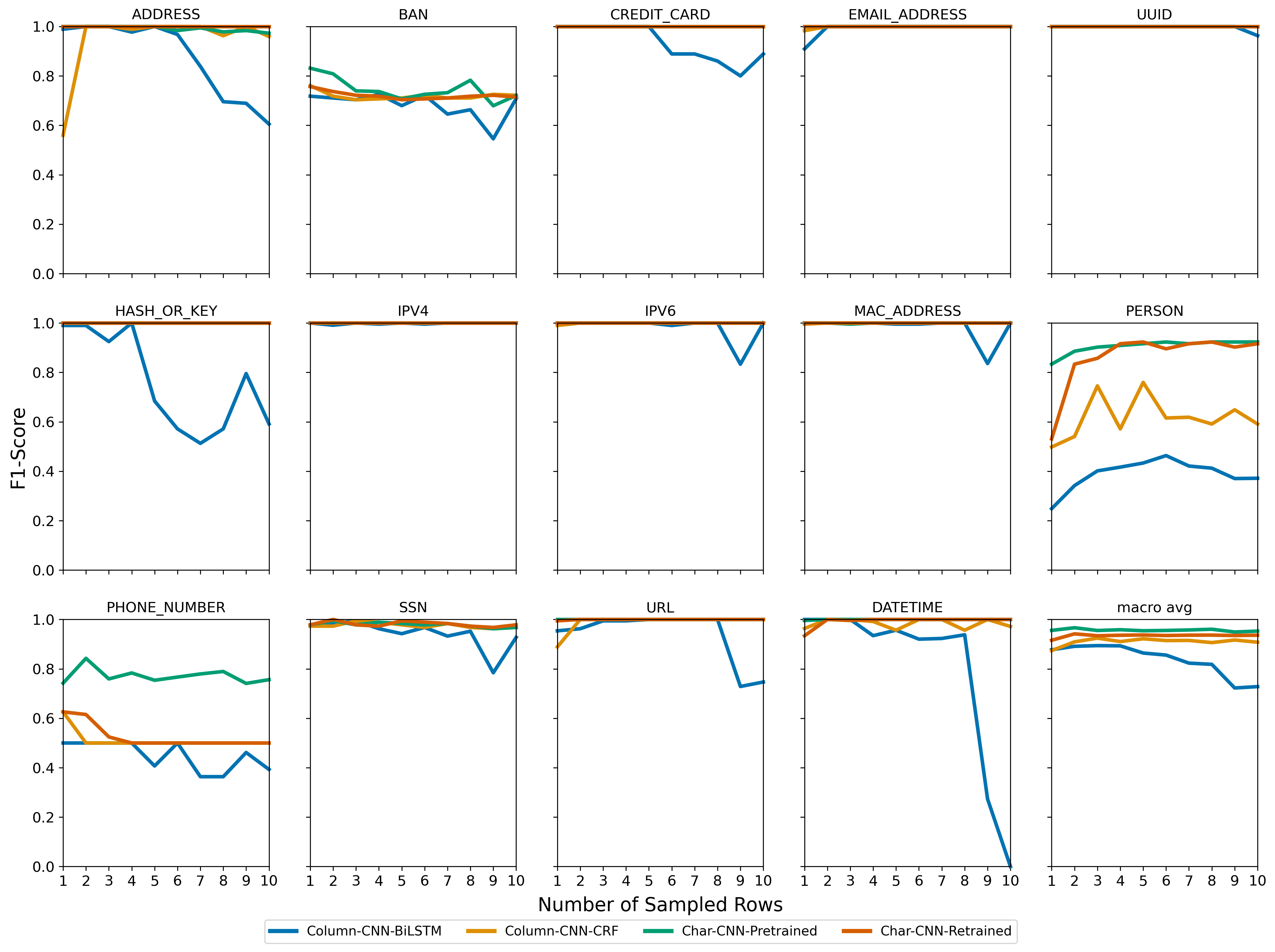}}
\caption{Accuracy breakdown at entity level for multi-column structured data. Empty subplots indicate the non-existence of the sensitive entities.}
\label{fig:acc-structured-entity-random-schema}
\end{center}
\end{figure*}

\begin{figure*}[!htb]
\begin{center}
\centerline{\includegraphics[width=1.5\columnwidth]{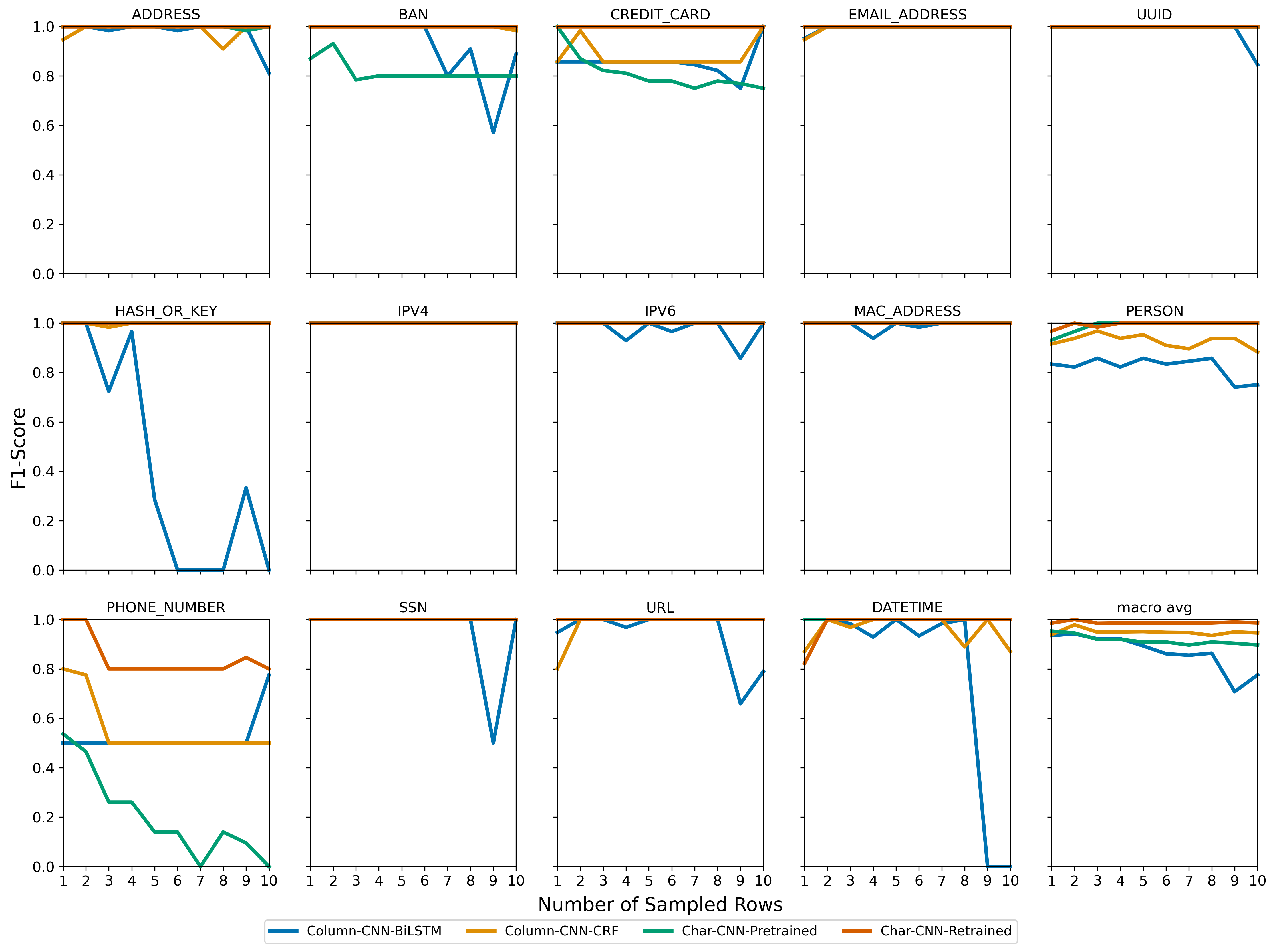}}
\caption{Accuracy breakdown at entity level for single-column structured data. Empty subplots indicate the non-existence of the sensitive entities.}
\label{fig:acc-structured-entity-one-column}
\end{center}
\end{figure*}

\begin{figure*}[!htb]
\begin{center}
\centerline{\includegraphics[width=1.5\columnwidth]{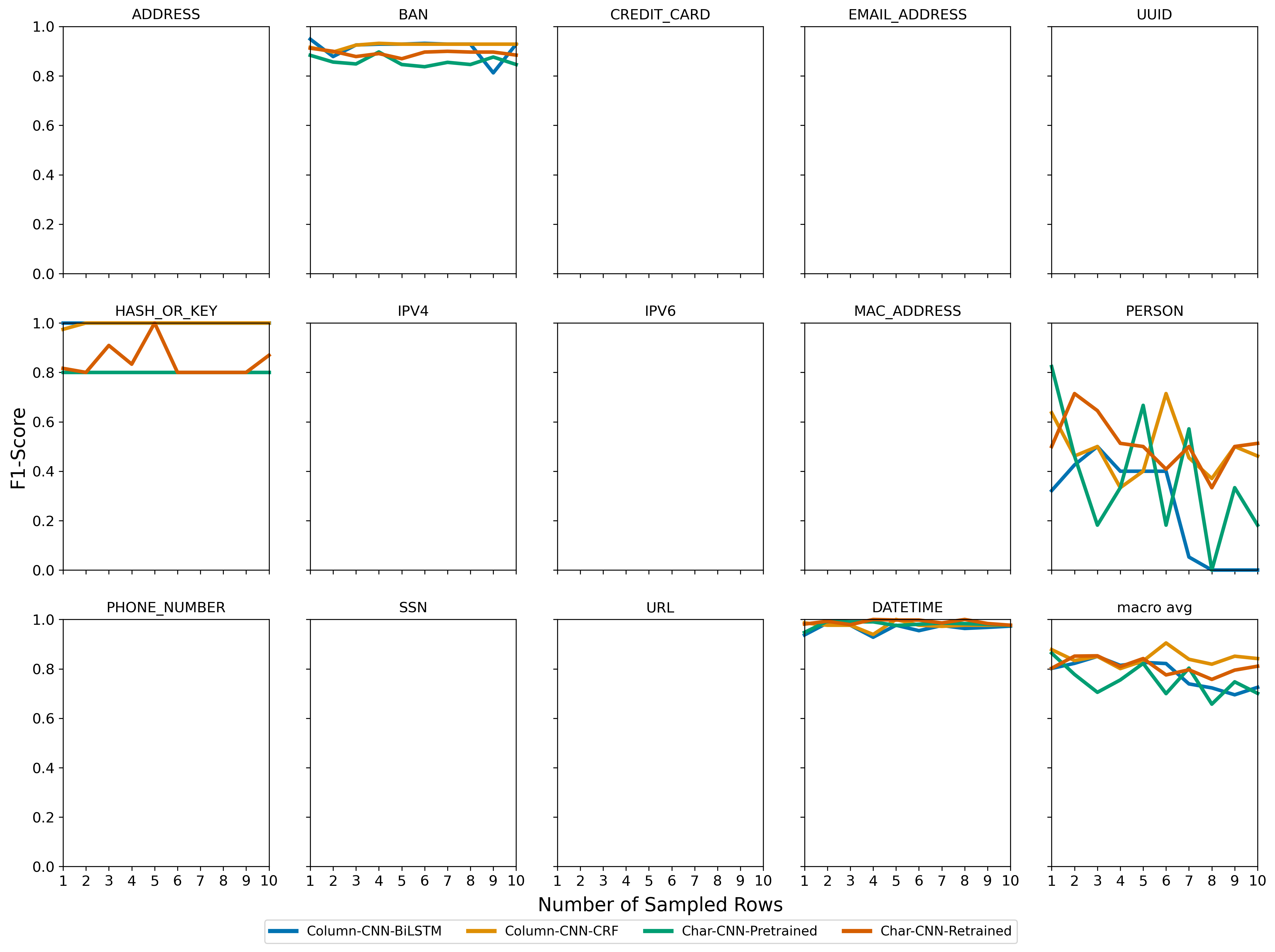}}
\caption{Accuracy breakdown at entity level for internal structured data. Empty subplots indicate the non-existence of the sensitive entities.}
\label{fig:acc-structured-entity-internal-data}
\end{center}
\end{figure*}

As with the entity detection task, a breakdown of accuracy results for individual sensitive entities in the multi-column structured data, single-column structured data, and internal structured data are given in Figure \ref{fig:acc-structured-entity-random-schema}, Figure \ref{fig:acc-structured-entity-one-column}, and Figure \ref{fig:acc-structured-entity-internal-data}, respectively.
It is shown from Figure \ref{fig:acc-structured-entity-internal-data} that BAN and PERSON gain higher accuracy as compared to the results of the entity detection task. This follows from the fact that aggregating samples for each entity with consistent formats reduces the confusion with other entities.
\end{document}